\documentclass[10pt,journal,compsoc]{IEEEtran}

\ifCLASSOPTIONcompsoc
  \usepackage[nocompress]{cite}
\else
  \usepackage{cite}
\fi

 \usepackage{graphicx}
\usepackage{booktabs}
\usepackage{threeparttable}
\usepackage{makecell}
\usepackage{wrapfig}
\usepackage{longtable}
\usepackage{caption}
\usepackage{subcaption}
\usepackage{array}
\usepackage{comment}
\usepackage{wrapfig}
\usepackage{longtable}
\usepackage{supertabular}
\usepackage[normalem]{ulem}
\usepackage{amsmath}
\usepackage{amssymb}
\DeclareTextFontCommand{\textbf}{\bfseries}
\DeclareTextFontCommand{\textit}{\itshape}

\usepackage{algorithmic}
\usepackage{algorithm}

\usepackage{subcaption} 
\usepackage{multirow} 

\ifCLASSINFOpdf
\else
\fi

\begin{document}

\title{ASP: Learn a Universal Neural Solver!}
%
%
%
%

\author{Chenguang Wang,
        Zhouliang Yu,
        Stephen McAleer,
        \\
        Tianshu Yu,
        and~Yaodong Yang
\IEEEcompsocitemizethanks{
\IEEEcompsocthanksitem Chenguang Wang, Zhouliang Yu and Tianshu Yu are with School of Data Science, the Chinese University of Hongkong, Shenzhen, China.
E-mail: \{chenguangwang, zhouliangyu\}@link.cuhk.edu.cn,  yutianshu@cuhk.edu.cn

\IEEEcompsocthanksitem Stephen McAleer is with Carnegie Mellon University,
USA.
E-mail: smcaleer@cs.cmu.edu

\IEEEcompsocthanksitem Yaodong Yang is with Institute for AI, Peking University, Beijing, China.\protect\\
E-mail: yaodong.yang@pku.edu.cn
}
}

\IEEEtitleabstractindextext{%
\begin{abstract}
Applying machine learning to combinatorial optimization problems has the potential to improve both efficiency and accuracy. However, existing learning-based solvers often struggle with generalization when faced with changes in problem distributions and scales. In this paper, we propose a new approach called ASP: \textbf{A}daptive \textbf{S}taircase \textbf{P}olicy Space Response Oracle to address these generalization issues and learn a universal neural solver. ASP consists of two components: Distributional Exploration, which enhances the solver's ability to handle unknown distributions using Policy Space Response Oracles, and Persistent Scale Adaption, which improves scalability through curriculum learning. We have tested ASP on several challenging COPs, including the traveling salesman problem, the vehicle routing problem, and the prize collecting TSP, as well as the real-world instances from TSPLib and CVRPLib. Our results show that even with the same model size and weak training signal, ASP can help neural solvers explore and adapt to unseen distributions and varying scales, achieving superior performance. In particular, compared with the same neural solvers under a standard training pipeline, ASP produces a remarkable decrease in terms of the optimality gap with 90.9\% and 47.43\% on generated instances and real-world instances for TSP, and a decrease of 19\% and 45.57\% for CVRP.
\end{abstract}

}

\maketitle

\IEEEdisplaynontitleabstractindextext

%
\IEEEpeerreviewmaketitle

\IEEEraisesectionheading{\section{Introduction}\label{sec:introduction}}

%
%
%
%

\IEEEPARstart{C}OMBINATORIAL optimization problems (COPs) have received significant attention from the operation research (OR) and theoretical computer science (TCS) communities due to their numerous applications. Both OR and TCS have developed a variety of solvers to tackle COPs, including both exact and heuristic approaches. However, the design of these solvers is extremely labor-intensive and requires extensive domain knowledge, making it difficult to create new solvers.

Recently, COPs have also attracted attention from the machine learning community, as it was discovered that deep learning-based "neural solvers" can capture complex structures and heuristics through the analysis of large numbers of problem instances~\cite{dai2017learning}. These neural solvers have been found to be very efficient on large-scale problems, and have the added benefit of relieving the tedious labor required to design traditional solvers. However, current neural solvers suffer from generalization issues in two main areas: problem distribution and scalability.

Almost all previous neural solvers have been trained and tested on a well-defined problem distribution and scale, and do not perform well when applied to real-world scenarios where the problem distribution and scale can vary significantly~\cite{dai2017learning,kool2018attention,wu2021learning,kool2021deep,kwon2020pomo}. This limits the use of neural solvers as alternatives or replacements for traditional solvers. One potential solution to the generalization issues faced by previous neural solvers is to train different solvers for different settings or to use a very large model that can handle a wide range of problem instances. However, these approaches have drawbacks. Training multiple solvers can be computationally intensive and require significant resources, while using a very large model can sacrifice efficiency and may result in degraded performance when adapting to different settings. As a result, we are interested in the following question:

\begin{itemize}
    \item \emph{Can we obtain a capacity-efficient neural solver without sacrificing performance while overcoming distributional and scale issues?}
\end{itemize}

\noindent{Several recent studies have attempted to address the generalization issues faced by previous neural solvers for COPs, but from specific angles. For example, some research has focused on creating new distributions during training to address the issue of data distribution~\cite{wang2021game, Zhang2022LearningTS}, while others have examined ways to adapt neural solvers to different scales~\cite{lisicki2020evaluating}. However, none of these methods offer a comprehensive, systematic approach to addressing the generalization issues universally.}

In this paper, we propose a new method called ASP: Adaptive Staircase Policy Space Response Oracle (PSRO), which aims to address the generalization issues faced by previous neural solvers and provide a "universal neural solver" that can be applied to a wide range of problem distributions and scales. Our method incorporates game theory and curriculum learning, and demonstrates improved performance on a variety of COPs compared to traditional solvers and previous neural solvers. Specifically, our method follows the two-player zero-sum meta-game framework proposed in \cite{wang2021game}, and employs policy space response oracles (PSROs) to adapt to different problem distributions and scales. We also introduce the concept of a staircase policy~\cite{treutwein1995adaptive}, which allows our method to learn multiple policies at once and make more efficient use of the model's capacity. 

ASP consists of two main components: Distributional Exploration (DE) and Persistent Scale Adaption (PSA). DE uses policy space response oracles (PSROs) to adapt the neural solver to different problem distributions, while PSA uses an adaptive staircase procedure to gradually increase the difficulty of the problem scale, allowing the neural solver to persistently improve its performance on a wide range of scales. Different from \cite{wang2021game} which generates the attack distribution by learning mixed Gaussian perturbations and approximating policy gradient via the Monte Carlo method, we utilize Real-NVP~\cite{dinh2016density}, a more generic generation approach via Normalizing Flow which can generate various data distributions with exact probability. 

We have tested ASP on several classical COPs, including the traveling salesman problem (TSP) and the capacitated vehicle routing problem (CVRP), using two typical RL-based neural solvers: Attention Model (AM)~\cite{kool2018attention} and POMO~\cite{kwon2020pomo}. Our results show that ASP can achieve impressive improvements compared to the original solvers trained under standard paradigms, and outperforms all other DL-based neural solvers on both TSPLib~\cite{reinelt1991tsplib} and CVRPLib~\cite{uchoa2017new} datasets. Additionally, we have conducted ablation studies to demonstrate the effectiveness of our approach and the influence of model capacity on training. 

In summary, our main contributions are as follows:
\begin{itemize}
    \item We propose the ASP framework, which is designed to overcome the generalization issues faced by previous neural solvers for COPs. This framework is model- and problem-agnostic, making it widely applicable to a range of neural solvers and COPs.
    \item Without increasing the model capacity, ASP allows a neural solver to discover complex structures by interchangeably learning from different distributions and scales, resulting in state-of-the-art performance even with comparable training resources to other methods trained under a fixed setting.
    \item We conduct extensive ablation studies and provide insights into training strategy design, model capacity, the function landscape of COPs, and beyond. These findings can broaden the horizons of future research in the field of COPs.

\end{itemize}

\section{Related Work}
\textbf{Deep Learning for Combinatorial Optimization Problems} -  
 Pointer Networks~\cite{vinyals2015pointer} is the first attempt to solve COPs with deep learning, specifically by training a pair of RNN-based encoder and decoder to output a permutation over its inputs. 
 \cite{bello2016neural} applies a similar model but instead trained the model with reinforcement learning rather than supervised learning used in Pointer Networks, by treating the tour length following the node sequence as the reward signal. 
 Inspired by the success of Transformers~\cite{vaswani2017attention}, Attention Model~\cite{kool2018attention} uses the attention mechanism, which is similar to Graph Attention Networks~\cite{velivckovic2017graph}, to encode the node representation and decode the solutions to various problems. \cite{lu2019learning} proposes the ``Learn to Improve'' (L2I) model which iteratively refines the solution with improvement operators, and jumps out of local optima with perturbation operators. 
 Similarly, \cite{wu2021learning} proposed a reinforcement learning framework to learn the improvement heuristics for TSP and the Capacitated Vehicle Routing Problem (CVRP) and achieves excellent results both on randomly generated instances and real-world instances. Some recent works~\cite{kool2021deep,fu2020generalize, agostinelli2019solving, agostinelli2021search} focus on pretraining models and constructing heatmaps to guide the search method.
 Although deep learning- or reinforcement learning-based methods have achieved notable progress in various COPs, in addition to generalization issues, some other essential limitations also exist, for example, degenerated performance on large-scale problems and high computational consumption during training. Finally, we refer to \cite{bengio2020machine} for a comprehensive review of the existing challenges in this area.
 
 \noindent\textbf{Generating Hard Instances} - There are some previous works that seek to obtain robust algorithms by generating hard-to-solve instances~\cite{smith2015generating, zuzic2020learning,liu2020generative}. \cite{smith2015generating} proposed a method to generate a representation of the problem instance space, with which the performance of certain algorithms can be quantified and thus the generalization can be inferred. Using the measurable features, they can analyze the similarities and differences between instances and the strengths and weaknesses of algorithms. Closely related to our work, \cite{zuzic2020learning} focuses on AdWords problem and constructs an adversarial training framework inspired by game theory, which employs GANs~\cite{goodfellow2014generative} to generate adversarial instances to expose the weakness of any given algorithms. \cite{liu2020generative} considers the instance generation and portfolio construction in an adversarial fashion to reduce the impact of deviation of training data. Different from these works, we adopt the PSRO~\cite{lanctot2017unified} framework to obtain a population of learnable solvers through meta-game and devise a model mixture method to combine these solvers, which leads to state-of-the-art performance. In future work we will investigate applying new PSRO variants for improved performance~\cite{mcaleer2022anytime, perez2021modelling, mcaleer2022self}. One could also potentially apply other algorithms for two-player zero sum games to our objective~\cite{perolat2022mastering, fu2021actor, mcaleer2022escher}.

\noindent\textbf{Curriculum Learning} - \cite{bengio2009curriculum} firstly propose the concept of curriculum learning (CL) where a machine learning algorithm should be trained from easy tasks to hard tasks. 
Being a plug-and-play module with high flexibility, curriculum learning has been broadly used in various research fields: computer vision~\cite{Pentina2015CurriculumLO, guo2018curriculumnet}, natural language processing \cite{tay2019simple,Platanios2019CompetencebasedCL}, reinforcement learning (RL) ~\cite{florensa2017reverse,ren2018self,Narvekar2020CurriculumLF,Portelas2020AutomaticCL,Huang2022CurriculumbasedAM}, to name a few. Numerous successful applications of CL have demonstrated its capability of improving the performance on target tasks and accelerating the convergence of training process. Specifically for RL, CL can improve sample efficiency and asymptotic performance which are considered beneficial for the exploration and generalization \cite{Portelas2020AutomaticCL}. Closely related to our work, \cite{Zhang2022LearningTS} introduces ``adaptive-hardness'' to assess the generalizability of the neural solver and proposes a curriculum learner to improve the neural solver's generalizability over data distribution. \cite{lisicki2020evaluating} instead utilizes curriculum learning to improve the scalability of neural solvers. But these two works are limited to only one dimension of generalization issues respectively: \cite{Zhang2022LearningTS} focuses on generating hard instances with respect to the data distribution; \cite{lisicki2020evaluating} cares more about the problem scale. In this work, we take these two dimensions into consideration together.

\section{Notations and Preliminaries}
\textbf{Normal Form Game (NFG)} - A tuple $(\Pi,\mathbf{U}^{\Pi},n)$ where $n$ is the number of players, $\Pi=(\Pi_1,\Pi_2,...,\Pi_n)$ is the joint policy set and $\mathbf{U}^{\Pi}=(\mathbf{U}^{\Pi}_1,\mathbf{U}^{\Pi}_2,...\mathbf{U}^{\Pi}_n):\Pi\rightarrow\mathbb{R}^n$ is the utility matrix for each joint policy. A game is symmetric if all players have the same policy set ($\Pi_i=\Pi_j,i\ne j$) and same payoff structures, such that players are interchangeable.

\noindent\textbf{Best Response} - The strategy which attains the best expected performance against a fixed opponent strategy. $\sigma_{i}^{*}=\text{br}(\Pi_{-i},\sigma_{-i})$ is the best response to $\sigma_{-i}$ if:
\begin{equation*}
    \mathbf{U}^{\Pi}_i(\sigma^{*}_i,\sigma_{-i})\ge \mathbf{U}^{\Pi}_i(\sigma_i,\sigma_{-i} ), \forall i,\sigma_i\ne \sigma^{*}_i
\end{equation*}

\noindent\textbf{Nash Equilibrium} - A strategy profile $\sigma^{*}=(\sigma^{*}_1,\sigma^{*}_2,...,\sigma^{*}_n)$ such that:
\begin{equation*}
\mathbf{U}^{\Pi}_i(\sigma^{*}_i,\sigma^{*}_{-i})\ge \mathbf{\Pi}_i(\sigma_i,\sigma^{*}_{-i} ), \forall i,\sigma_i\ne \sigma^{*}_i
\end{equation*}
Intuitively, no player has an incentive to deviate from their current strategy if all players are playing their respective Nash equilibrium strategy.

\noindent\textbf{Traveling Salesman Problem (TSP)} - The objective is to find the shortest route that visits each location only once and return to the original location. In this paper, we only consider the two-dimension euclidean case, that is, each location's information comprises $(x_i,y_i)\in\mathbb{R}^{2}$.

\noindent\textbf{Vehicle Routing Problem (VRP)} - In the Capacitated VRP (CVRP)~\cite{toth2014vehicle}, there is a depot node and several demand nodes, the vehicle starts and ends at the deport node in multiple routes where the total demand of the demand nodes in each route does not exceed the vehicle capacity. The objective of the CVRP is to minimize the total route cost while satisfying all the constraints. We also consider the Split Delivery VRP (SDVRP), which allows to split customer demands over multiple routes.


\noindent\textbf{Prize Collecting TSP (PCTSP)} - 
In the PCTSP~\cite{balas1989prize}, the salesman who travels the given two-dimension locations, gets a prize in each location that he visits and pays a penalty to those that he fails to visit. The objective is to minimize the traveling length and penalties, while visiting enough cities to collect a prescribed amount of prizes. We also consider the Stochastic PCTSP (SPCTSP) in which the expected location prize is known upfront, but the real collected prize only becomes known upon visitation.

\noindent\textbf{Instance} - An individual sample of a combinatorial optimization problem. For example, given the two-dimensional coordinates of $n$ points, finding the shortest tour that traverses all points is an instance of TSP. Hereafter, we denote an instance with problem scale $n$ by $\mathcal{I}^n$ which comes from some distribution $\mathbf{P}_{\mathcal{I}^n}$.

\noindent\textbf{Optimality gap} - Measures the quality of a solver compared to an optimal oracle. Given an instance $\mathcal{I}^n$ and a solver $S:\{\mathcal{I}^n\}\rightarrow\mathbb{R}$, the optimality gap is defined as:
\begin{equation}\label{og}
    g(S,\mathcal{I}^n, \text{Oracle})=\frac{S(\mathcal{I}^n)-\text{Oracle}(\mathcal{I}^n)}{\text{Oracle}(\mathcal{I}^n)}
\end{equation}
where Oracle$(\mathcal{I}^n)$ gives the true optimal value of the instance. Furthermore, the \textit{expected} optimality gap of solver under an instance distribution $\mathbf{P}_{\mathcal{I}^n}$ is defined as:
\begin{equation}\label{eog}
    G(S,\mathbf{P}_{\mathcal{I}^n},\text{Oracle})=\mathbf{E}_{\mathcal{I}^n\sim\mathbf{P}_{\mathcal{I}^n}}g(S,\mathcal{I}^n,\text{Oracle}).
\end{equation}

\section{Method}
\begin{figure*}
    \centering
    \includegraphics[width=.8\textwidth]{  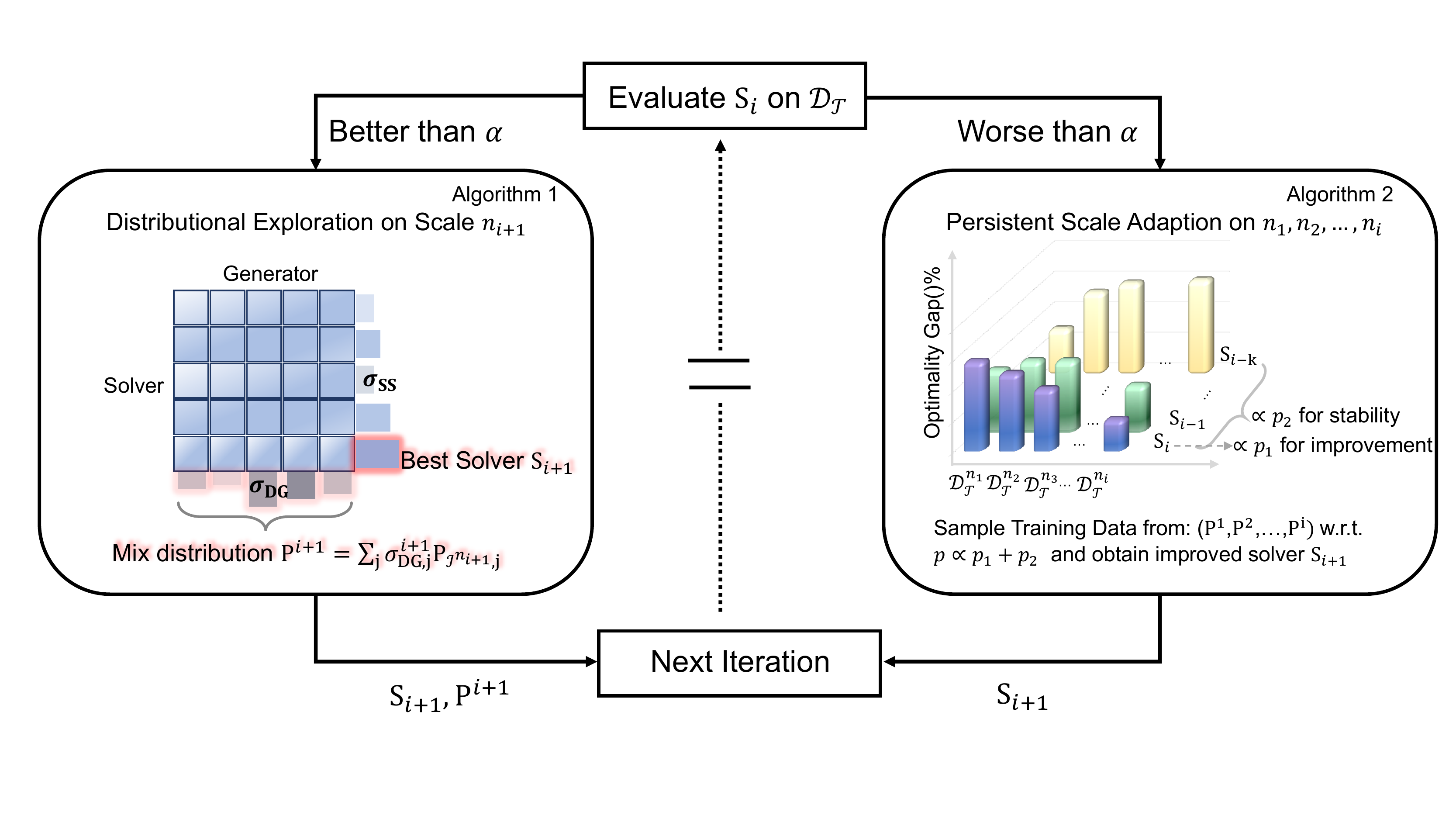}
    \vspace{-30pt}
    \caption{The pipeline of ASP framework. At the $i$-th iteration, ASP chooses to either improve the generalization ability on a larger problem scale $n_{i+1}$ or enhance the generalization ability on a set of pre-selected problem scales, depending on whether the performance of the current neural solver on the evaluation set is better than the preset threshold $\alpha$. If the performance is better than $\alpha$, we conduct the DE procedure on scale $n_{i+1}$, as detailed in algorithm \ref{alg:PSRO}, obtaining the best solver $S_{i+1}$ and mixed data distribution $\text{P}^{i+1}$ according to the meta-strategy. If not, we turn to improve $S_i$'s ability with the PSA procedure by training on problem scales $n_1,n_2,...,n_i$, as detailed in algorithm \ref{alg:task selection}. To sample training data from $\text{P}^j$ under problem scale $n_j$, we propose a sample strategy (Eq. \ref{eq: task selection obj}) based on the current and historical performance on evaluation set which balances the tendency of improvement and stability. Then an enhanced solver $S_{i+1}$ is obtained for the next iteration.}
    \label{fig:pipeline}
\end{figure*}
In this section, we present the framework, {ASP}, to obtain a universal neural solver with strong generalization ability on distribution and scale. In general, {ASP} contains two major components: \textbf{(1)} Distributional Exploration (DE): PSRO-based training for the improvement over distribution changes under a specific problem scale; and \textbf{(2)} Persistent Scale Adaption (PSA): Adaptive Staircase based curriculum for enhancing the generalization on a range of problem scales. The overview of ASP is shown in Fig.~\ref{fig:pipeline}. In the following parts, we will elaborate on them in detail.

\subsection{Distributional Exploration}\label{psro sec}
\begin{algorithm}[t]
\caption{Distributional Exploration (\textbf{DE}) with PSRO}
  \label{alg:PSRO}
\begin{algorithmic}
  \STATE \textbf{Input:} Neural Solver $S$, problem scale $n$
  \STATE \textbf{Initialize:} Joint policy set $\Pi=\{(S^n,\mathbf{P}_{\mathcal{I}^n})\}$, utility matrix $\mathbf{U}^{\Pi}$, meta-strategies $\sigma=(\sigma^n_{\text{SS}},\sigma_{\text{DG}})$
  \WHILE{epoch e in $\{1,2,...\}$}
        \STATE Train Oralces: $(S^{'}, \mathbf{P}^{'}_{\mathcal{I}^n})$ under problem scale $n$ w.r.t. Eq.~\ref{gradient of LSS} and Eq.~\ref{gradient of attack1 TSP}
    \STATE Update policy set: $\Pi\leftarrow\Pi\cup\{(S^{'},\mathbf{P}^{'}_{\mathcal{I}^n})\}$
    \STATE Update $\mathbf{U}^{\Pi}$ from new $\Pi$ and compute the new meta-strategy $\sigma$
    \ENDWHILE
    \STATE {\bfseries Output:} The best solver $S^n_{\text{best}}$ chosen from $\Pi_{\text{SS}}$ and mixed distribution $\mathbf{P^n}=\sum_i \sigma_{\text{DG},i}\mathbf{P}_{\mathcal{I}^n,i}$
\end{algorithmic}
\end{algorithm}

In this section, we formulate Distributional Exploration as a two-player game at the meta-level to improve the neural solver's generalization ability on the distribution dimension. When handling COP instances with problem scale $n$, we assume there are two players in the meta-game:
\begin{itemize}
    \item $\Pi_{\text{SS}}=\{S_i^n\mid i=1,2,...\}$ is the policy set for the Solver Selector where $S_i^n$ is the neural solver;
    \item $\Pi_{\text{DG}}=\{\mathbf{P}_{\mathcal{I}^n,i},i=1,2,...\}$  is the instance distribution policy set where $\mathbf{P}_{\mathcal{I}^n,i}$ is the instance distribution under problem scale $n$.
\end{itemize}
Formally, we have a two player asymmetric NFG $(\Pi, \mathbf{U}^{{\Pi}}, 2)$,
$\Pi=(\Pi_{\text{SS}},\Pi_{{\text{DG}}})$ is the joint policy set and $\mathbf{U}^{\Pi}:\Pi\rightarrow\mathbb{R}^{\mid\Pi_{\text{SS}}\mid\times \mid\Pi_{\text{DG}}\mid}$ is the utility matrix. Given a joint policy $\pi=(S^n,\mathbf{P}_{\mathcal{I}^n})\in\Pi$, its utility is given by
$$\mathbf{U}^{\Pi}(\pi)=\big(\mathbf{U}^{\Pi_{\text{SS}}}(\pi),\mathbf{U}^{\Pi_{\text{DG}}}(\pi)\big)$$
where $\mathbf{U}^{\Pi_{\text{SS}}}(\pi)=-\mathbf{U}^{\Pi_{\text{DG}}}(\pi)=-G(\pi,\text{Oracle})$ is the expected optimality gap under the joint policy $\pi$ as defined in Eq.~\ref{eog}.
The overall objective in the meta-level is:
\begin{equation}\label{eq:game formulation}
    \min_{\sigma_{\text{SS}}\in\Delta(\Pi^n_{\text{SS}})}\max_{\sigma_{\text{DG}}\in\Delta(\Pi^n_{\text{DG}})}\mathrm{E}_{\pi\sim(\sigma_{\text{SS}},\sigma_{\text{DG}})}G\big(\pi,\text{Oracle}\big).
\end{equation}
where $\Delta(\cdot)$ denotes the distribution on the given set.
We obey the PSRO framework as follows: at each iteration given the policy sets $\Pi=(\Pi_{\text{SS}},\Pi_{{\text{DG}}})$ and the meta-strategy $\sigma=(\sigma_{\text{SS}},\sigma_{\text{DG}})$, we train two Oracles:
\begin{itemize}
    \item $S^{'}$ represents a new neural solver, which is the best response to the meta-strategy $\sigma_{\text{DG}}$.
    \item $\mathbf{P}_{\mathcal{I}^n}^{'}$ represents a new instance distribution where Solver Selector performs poorly when following $\sigma_{\text{SS}}$.
\end{itemize}
 Given these oracles, we update the joint policy set $\Pi\leftarrow\Pi\cup(S^{'},\mathbf{P}_{\mathcal{I}_n}^{'})$ and the meta-game $\mathbf{U}^{\Pi}$ according to new $\Pi$. This expansion of the joint policy set has a dual purpose in that it aids in finding the difficult-to-solve instance distributions whilst also improving the ability of the Solver Selector to handle them. 
The general algorithm framework can be seen in Alg.~\ref{alg:PSRO}.

The formulation above leaves us with three algorithmic components to address:  \textbf{(1)} How to obtain the meta-strategy $\sigma$; \textbf{(2)} How to train Oracles for each player; \textbf{(3)} How to evaluate the utilities $\mathbf{U}^{\Pi}$. In the following parts, we will describe the flow of the algorithm.

\noindent \textbf{Meta-Strategy Solvers -} Given that we formulate a two-player game in Eq.~\ref{eq:game formulation}, it's preferable to use Nash Equilibrium of the meta-game as the meta-strategy solution due to its computational efficiency.

\noindent \textbf{Oracle Training -} We now provide the higher-level results for training a best-response Oracle, with more detailed derivations presented in Appendix~\ref{app:derive gradient}. Here we represent the neural solvers in $\Pi_{\text{SS}}$ as $S_{\theta}$ and the instance distributions in $\Pi_{\text{DG}}$ as $\mathbf{P}_{\mathcal{I}^n,\gamma}$ where $\theta$ and $\gamma$ are the trainable parameters.
\begin{itemize}
    \item For solver Oracle: Given the meta-strategy $\sigma_{\text{DG}}$, the oracle training objective for the neural solver is:
\begin{equation}
    \min_{\theta}L_{\text{SS}}(\theta)=\mathbf{E}_{\mathbf{P}_{\mathcal{I}^n}\sim\sigma_{\text{DG}}}G(S_\theta,\mathbf{P}_{\mathcal{I}^n},\text{Oracle}).
\end{equation}
We can optimize this objective by stochastic gradient descent and the gradient is:
\begin{equation}\label{gradient of LSS}
\begin{aligned}
&\nabla_{\theta}L_{\text{SS}}(\theta)
    &=\mathbf{E}_{\mathbf{P}_{\mathcal{I}^n}\sim\sigma_{\text{DG}}}\mathbf{E}_{{\mathcal{I}^n}\sim\mathbf{P}_{\mathcal{I}^n}}\frac{\nabla_{\theta}S_\theta(\mathcal{I}^n)}{\text{Oracle}(\mathcal{I}^n)}.
\end{aligned} 
\end{equation}
Given that it's impossible to call ``Oracle'' in Eq.~\ref{gradient of LSS} for excessively numerous iterations during training, we can alternatively train solver Oracle to obtain a powerful neural solver given distribution $\mathbf{P}_{\mathcal{I}}$ with the gradient:
\begin{equation}\label{LSS gradient during implement}
\begin{aligned}
&\nabla_{\theta}L_{\text{SS}}(\theta)
    &=\mathbf{E}_{\mathbf{P}_{\mathcal{I}^n}\sim\sigma_{\text{DG}}}\mathbf{E}_{{\mathcal{I}^n}\sim\mathbf{P}_{\mathcal{I}^n}}\nabla_{\theta}S_\theta(x_1,...,x_n).
\end{aligned} 
\end{equation}

\item For data generator Oracle: Given the meta-strategy $\sigma_{\text{SS}}$, the oracle training objective for the DG is:
\begin{equation}
    \max_{\gamma}L_{\text{DG}}(\gamma)=\mathbf{E}_{S\sim\sigma_{\text{SS}}}G(S,\mathbf{P}_{\mathcal{I}^n,\gamma},\text{Oracle}).
\end{equation}

The gradient w.r.t. $\gamma$ is:
\begin{equation}\label{gradient of attack1 TSP}
\begin{aligned}
    &\nabla_{\gamma_C}L_{\text{DG}}(\gamma)
    =\mathbf{E}_{S^n\sim\sigma_{\text{SS}}}\mathbf{E}_{\mathcal{I}^n\sim\mathbf{P}_{\mathcal{I}^n,\gamma}}\Big[\\
    & \qquad \qquad \nabla_{\gamma}\big(\log\mathbf{P}_{\mathcal{I}^n,\gamma}(\mathcal{I}^n)\big)g\big(S,\mathcal{I}^n,\text{Oracle}\big) \Big].
\end{aligned}
\end{equation}
Similar with the training oracles for the neural solver, we can omit the call of the Oracle solver as well.
Otherwise, to accurately calculate the probability $\mathbf{P}_{\mathcal{I}^n,\gamma}(\mathcal{I}^n)$ in Eq.~\ref{gradient of attack1 TSP}, we apply Real-NVP~\cite{dinh2016density}, a Normalizing Flow-based generative model to generate adversarial data. Concretely, we start from a simple prior probability distribution
\begin{equation}
    z\sim p_Z = Uni.([0,1]^k)
\end{equation}
with the probability $p_Z(z)=1,\forall z\in[0,1]^k$ where $k$ is the dimension of input feature. We obtain the adversarial data by a parameterized bijection 
$$x = G_\gamma (z):Z\rightarrow X,$$ 
then the exact probability of obtaining $x$ is given by the change of variable formula:
$$
p_X(x) = p_Z(G^{-1}(x))\big|\det \big(\frac{\partial G^{-1}_\gamma}{\partial x}\big)\big|=\big|\det \big(\frac{\partial G^{-1}_\gamma}{\partial x}\big)\big|.
$$
This probability can be computed efficiently and exactly thanks to the construction of $f_\gamma$ in Real-NVP, so we are able to calculate the probability $\mathbf{P}_{\mathcal{I}^n,\gamma}(\mathcal{I}^n)$ in this way.
\end{itemize}

\noindent \textbf{Evaluation -} Given the joint policy $\pi\in\Pi$, we can compute the element in the utility matrix $\mathbf{U}^{\Pi}(\pi)=\big(-G(S,\mathbf{P}_{\mathcal{I}^n},\text{Oracle}),G(S,\mathbf{P}_{\mathcal{I}^n},\text{Oracle})\big)$ by approximating the expected optimality gap defined in Eq.~\ref{eog}:
\begin{equation}\label{eq:evaluation}
\begin{aligned}
        G(S,\mathbf{P}_{\mathcal{I}^n},\text{Oracle})
    &=\mathbf{E}_{\mathcal{I}^n\sim\mathbf{P}_{\mathcal{I}^n}}g(S,\mathcal{I}^n,\text{Oracle})\\
    &\approx\frac{1}{M}\sum_{i=1}^{M}g(S,\mathcal{I}^n_i,\text{Oracle}).
\end{aligned}
\end{equation}
where $S\in\Pi_{\text{SS}}, \mathbf{P}_{\mathcal{I}^n}\in\Pi_{\text{DG}}$.

After finishing the training of PSRO for problem scale $n$, we obtain the best neural solver $S^n_{\text{best}}$ and a mixed distribution $\mathbf{P^n}=\sum_i \sigma_{\text{DG},i}\mathbf{P}_{\mathcal{I}^n,i}$ for later usage.

\subsection{Persistent Scale Adaption}\label{sec: PSA}
\begin{algorithm}[H]
\caption{Task Selection by Persistent Scale Adaption (\textbf{PSA})}
  \label{alg:task selection}
\begin{algorithmic}
  \STATE {\bfseries Input:} Neural Solver $S$, a range of problem scales $n_1,...,n_i$ and mixed distributions $\mathbf{P^{1}},...,\mathbf{P^{i}}$
  \STATE {\bfseries Initialize:} $p_0 \in \mathbb{R}^i$ with a uniform probability vector
  \WHILE{epoch t in $\{1,2,...\}$}
        \STATE Evaluate $S$ on each $(\mathbf{P^{k}},n_k)$ by Eq. \ref{eq:evaluation} to obtain cost vector $c^t=(c_{k}^t)_k$
        \STATE Compute the task selection strategy $p_t$ followed by Eq. \ref{eq: task selection obj} and train a new neural solver $S'$ w.r.t. Eq. \ref{eq:PSA} with $P_{\text{PS}}=p_t$
    \STATE Assign $S\leftarrow S^{'}$
    \ENDWHILE
    \STATE {\bfseries Output:} The new trained Neural Solver $S$.
\end{algorithmic}
\end{algorithm}

 Considering a range of problem scales $(n_1,n_2,...,n_i)$, 
 we want to obtain a universal neural solver which has persistent performance on all problem scales. In this section, we propose the Persistent Scale Adaption (PSA) procedure to achieve this under the framework of curriculum learning.
Concretely, we can formulate this as a multi-objective problem:
\begin{equation*}
        \min_{\theta}L^{M}_{\text{PSA}}(\theta)=\big(G(S_\theta,\mathbf{P}^{n_1},\text{Oracle}),...,G(S_\theta,\mathbf{P}^{n_i},\text{Oracle})\big).
\end{equation*}
A common way to optimize this is transferring it to a single objective problem by assigning each objective a weight, leading to the expectation form:
\begin{equation}\label{eq:PSA}
        \min_{\theta}L_{\text{PSA}}(\theta)=\mathbf{E}_{n\sim P_\text{PS}}G(S_\theta,\mathbf{P}^n,\text{Oracle}).
\end{equation}
where $P_\text{PS}$ is the task selection strategy, a discrete probability distribution over the problem scales $(n_1,n_2,...,n_i)$ such that $P_\text{PS}(n=n_k)=p_k,\ \sum_k p_k=1$. 
There are many candidates for $P_\text{PS}$, e.g. the uniform distribution as in the original Adaptive Staircase Procedure, but this would mislead the neural solver training on improper tasks. Here we propose a \textit{momentum-based} mechanism to guide the neural solver to focus on the proper tasks. 

We treat the sampling probability $\{p_i\}$ as a weak signal to guide training process towards the improvement on target distribution $\mathcal{D}_{\mathcal{T}}$. By evaluating the neural solver $S_\theta$ on each pair of $(\mathcal{D}_{\mathcal{T}},n_k)$ we can obtain a cost vector $c\in\mathbb{R}^i$:
$$
c=(c_{k})_k,\text{where }c_{k}\text{ is evaluated by Eq. \ref{eq:evaluation},}
$$
measuring the performance of $S_\theta$ on the data distribution $\mathcal{D}_{\mathcal{T}}$ under the problem scale $n_k$. 
During the implementation, we will execute the PSA procedure until reaching some conditions (e.g., achieving the pre-set performance or conducting the training for certain iterations), so we denote the cost vector at the $t$-th PSA iteration as $c^t=(c_{k}^t)_k$ and construct $\{p_i^t\}$ as follows:
\begin{equation}
\label{eq: task selection obj}
       p^t_i\propto \lambda p^1_i + (1-\lambda)p^2_i, \lambda\in[0,1]
\end{equation}
where $ p^1_i\propto \lfloor c^t_i - c^{t-1}_i\rfloor_+$ with $\lfloor x \rfloor_+ = x$ if $x>0$, otherwise $0$; $p^2_i\propto\sqrt{Var_t[c^t_i]}$. $p^1_i$ helps us focus on the problem scales on which the neural solver performs poorly and $p^2_i$ measures the stability over history performance, which benefits the training process with more stability. $\lambda$ is the weight to balance the two terms.

Since $c^t$ is unknown beforehand and changes over training time, we maintain a history of the previous 10 steps of cost vectors, and update $Var_t[c^t_i]$ dynamically during training, so that we can guide the training process to focus on those tasks on which the performance is unsatisfactory and unstable. At the beginning of training, we sample $n_i$ uniformly.
The whole process can be seen in Alg. \ref{alg:task selection}.

\subsection{The Whole ASP Pipeline}\label{curriculum sec}
\begin{algorithm}[hbt!]
\caption{\textbf{ASP} (\textbf{A}daptive-\textbf{S}tairs \textbf{P}SRO)}
  \label{alg:ASP}
\begin{algorithmic}
  \STATE \textbf{Input:} Neural Solver $S$, minimal and maximal problem scale $n_1, n_K$, incremental step $n_{\text{step}}$, evaluation dataset $D=\{\mathcal{D}_{\mathcal{T}}^{{n_1}},\mathcal{D}_{\mathcal{T}}^{{n_1+n_{\text{step}}}},...,\mathcal{D}_{\mathcal{T}}^{{n_K}}\}$ and performance threshold $\alpha$, training patience $\gamma$
    \STATE \textbf{Initialize:} Problem scale list $p_{l}=[n_1]$
     $S_{\text{best}},\ \mathbf{P^{1}}$ = \textbf{DE}($S,n_1$) by Alg. \ref{alg:PSRO}, mix distribution list $p_{\text{dist}}=[\mathbf{P^{1}}]$, count$=0$, evaluate $S_{\text{best}}$ on $\mathcal{D}_{\mathcal{T}}^{{n_1}}$ to get evaluation result $r$
  \WHILE{True}
  \IF{($r\le\alpha$ or count$>\gamma$) and $p_{l}[-1] \le n_K$} 
    \STATE $n{'}=p_{l}[-1]+n_{\text{step}}$ 
    \IF{$n{'}<n_K$}
        \STATE $p_{l}\leftarrow p_{l} + [n{'}]$
    \ELSE
        \STATE Break
    \ENDIF
    \STATE $S_{\text{best}},\mathbf{P^{\text{new}}}$ = \textbf{DE}($S_{\text{best}},p_{l}[-1]$) by Alg. \ref{alg:PSRO} 
    \STATE $p_{\text{dist}}\leftarrow p_{\text{dist}} + [\mathbf{P^{\text{new}}}]$
    \STATE count=0
  \ELSE 
  \STATE $S_{\text{best}}$ = \textbf{PSA}($S_{\text{best}},p_{l},p_{\text{dist}}$) by Alg. \ref{alg:task selection}
    \STATE count+=1
  \ENDIF
 \STATE Evaluate $S_{\text{best}}$ on $\{\mathcal{D}_{\mathcal{T}}^{{n_1}},...,\mathcal{D}_{\mathcal{T}}^{p_{l}[-1]}\}$ to get evaluation result $r$ 
    \ENDWHILE
    \STATE {\bfseries Output:} The best Neural Solver $S_{\text{best}}$
\end{algorithmic}
\end{algorithm}
In this section, we provide the overall details of the Adaptive Staircase PSRO (ASP) framework by integrating the components in previous sections.

Borrowing the idea from curriculum learning that starting from easy tasks and raising the difficulty level gradually, we begin to train the neural solver from the smallest problem scale because problems with larger scales are considered to be harder than smaller ones in the context of COPs. Inspired by Adaptive Staircase Procedure~\cite{treutwein1995adaptive}, at each ASP iteration, we consider two optional procedures: DE procedure described in algorithm \ref{alg:PSRO} and PSA procedure described in algorithm \ref{alg:task selection}. Given the problem scales we have handled $(n_1,n_2,...,n_i)$ and neural solver $S_\theta$ trained so far, if the evaluation results of $S_\theta$ on target distribution $\mathcal{D}_{\mathcal{T}}$ is better than a preset threshold $\alpha$, we will conduct DE procedure on a larger problem scale $n_{i+1}$, meaning taking a higher staircase to a harder task, otherwise we invoke PSA procedure on $(n_1,n_2,...,n_i)$, meaning enhancing $S_\theta$'s ability on current tasks. We repeat this process until some stop criterion is satisfied and the whole pipeline is shown in Alg. \ref{alg:ASP}.

\section{Experiments}
In this section, we present our results on several COPs: TSP, CVRP, SDVRP, PCTSP, and SPCTSP. 
 In contrast to previous work which fails to show generalization ability due to training and testing on the same distribution (uniform) and training a model for each problem scale, we demonstrate performance on distributions that are \emph{never trained} and evaluate \emph{only one} model on various problem scales. 

\subsection{Experimental Settings}\label{sec: exp setting}

\noindent\textbf{Base Solver -} Our method is model- and problem-agnostic so we employ existing neural solvers on various COPs to demonstrate the validity of this framework. We call the neural solver without the training of our method as \emph{base solver} and denote the neural solver using our method with the annotation \emph{base solver}(ASP).
\begin{table}[hbpt!]
\centering
\begin{tabular}{lll}
\multicolumn{1}{c}{\bf COP}  &\multicolumn{1}{c}{\bf Base Solver}  &\multicolumn{1}{c}{\bf Threshold}
\\ \midrule
\multirow{2}{*}{TSP}         &AM~\cite{kool2018attention} &3\\                                       &POMO~\cite{kwon2020pomo} &0.5\\

\multirow{2}{*}{CVRP}         &AM~\cite{kool2018attention}&5\\     
&POMO~\cite{kwon2020pomo}&1\\
SDVRP             &AM~\cite{kool2018attention} &10\\
PCTSP             &AM~\cite{kool2018attention} &10\\
SPCTSP             &AM~\cite{kool2018attention} &10\\
\midrule
\end{tabular}
\caption{Base solver and threshold for each COP}
\label{base solver}
\end{table}

\noindent\textbf{Training Setting -} 
All the training under different base solvers and COPs is from scratch.
Considering the different performance of neural solvers on different COPs, the choices of base solvers for each COP and settings of performance threshold $\alpha$ are different as well, as shown in Table \ref{base solver}. For the weight in Eq. \ref{eq: task selection obj}, we use $\lambda=0.5$ for all experiments.
We set the minimal and maximal problem scale $n_1=20, n_K=100$, and incremental step $n_{\text{step}}=20$ for all COPs.
During the training process, we evaluate the neural solver on target distribution $\mathcal{D}_{\mathcal{T}}$ under the problem scales that have never been trained and get average evaluation result $r$. A patience parameter $\gamma=5$ is also set to prevent the neural solver from getting stuck in the current tasks, that is, we will step to a harder task if no further improvements are observed after 5 training iterations of the PSA procedure.

The Data Generator is a Normalizing Flow-based model, Real-NVP~\cite{dinh2016density} with 5 coupling layers and we train its parameters using Adam optimizer~\cite{kingma2014adam} with a learning rate of 1e-4 for 100 epochs. What's more, we reset the parameters when we begin to conduct DE training for a larger problem scale. We leave the detailed settings about model structure and training configurations for base solvers and data generator in Appendix~\ref{app:details on dg}

\noindent\textbf{Baselines --} For TSP, we compare the solution quality and inference speed of the neural solvers trained by our framework with exact solvers, non-learning-based heuristics methods, and learning-based methods. More specifically, the counterpart solvers include Concorde, Gurobi~\cite{gurobi}, LKH3~\cite{helsgaun2017extension} which is a state-of-the-art heuristic solver, and OR-tools which is an approximate solver based on meta-heuristics. Moreover, we also compare against Nearest, Random, and Farthest Insertion. Additionally, we compare our methods with a variety of learning-based methods: LIH~\cite{lu2019learning}, AM~\cite{kool2018attention}, MDAM~\cite{xin2021multi}, POMO~\cite{kwon2020pomo}, GCN \cite{joshi2019efficient} and CVAE-opt \cite{hottung2020learning}. 
 For CVRP, we test the performance of the neural solvers trained by ASP in comparison with the following solvers: LKH3, and Gurobi with various time limits, as well as the following mentioned learning-based methods: AM, POMO, RL(gr.)~\cite{nazari2018reinforcement},  NeuRewriter~\cite{chen2019learning} and CVAE-opt.
 For SDVRP, we compare against RL(gr.), AM~\cite{kool2018attention}, and MDAM \cite{xin2021multi}. 
For PCTSP, we compare against Gurobi, OR-Tools, as well as the python version of Iterated Local Search(ILS). Moreover, we also compare our neural solver AM(ASP) to its baseline, AM~\cite{kool2018attention}. 
For SPCTSP, We compare our neural solver AM(ASP) against its baseline AM and MDAM.

\noindent\textbf{Time Consumption --} Training time is an essential facet of neural solvers. When applying ASP, there will be extra time consumption for training the data generator and evaluating the meta-payoffs in DE on top of the standard training consumption of neural solvers.
All the experiments are conducted on a single GeForce RTX 3090. Under our setting, training POMO on TSP and CVRP takes 1.67 days and 2.25 days, respectively; training AM on TSP and CVRP takes 4 days and 3.3 days, respectively. When training AM for other COPs, we stop the training early once there are no further improvements, resulting in a training duration of 2 days approximately. Compared with the training cost of base solvers, training a neural solver under ASP takes slightly more time than that base solver takes on a problem scale of 50, and much less than a problem scale of 100.

\subsection{Results on Generated Data}
\begin{table*}[h!]
\caption{Our model vs baselines. The gap \% is w.r.t. the best value across all methods. The bold numbers indicate base solver(ASP) achieves better performance than base solver and the underline numbers mean the best performance among all heuristic methods. We especially focus on the Avg. Gap because we care more about the performance of difference methods on several problem scales.}
\label{tab:results on generated data}
\centering
\renewcommand{\arraystretch}{0.9}
\scalebox{1.25}{
\begin{tabular}{ll|rr|rr|rr|r}
 & &  \multicolumn{2}{c|}{$n = 20$}  & \multicolumn{2}{c|}{$n = 50$} & \multicolumn{2}{c|}{$n = 100$} \\
 & Method 
 & \multicolumn{1}{c}{Gap} & \multicolumn{1}{c|}{Time}  
 & \multicolumn{1}{c}{Gap} & \multicolumn{1}{c|}{Time} 
 &  \multicolumn{1}{c}{Gap} & \multicolumn{1}{c|}{Time} 
 & \multicolumn{1}{c}{Avg. Gap}\\
\midrule
\midrule
\multirow{19}{*}{\rotatebox[origin=c]{90}{TSP}}
 &  Concorde  &  $0.00\%$ & $(18.50s)$   &  $0.00\%$ &  $(46.38s)$ & $0.00\%$  & $(34.52s)$ & $0.00\%$\\
 &  LKH3  &  $0.00\%$ & $(3.01s)$  & $0.00 \%$ & $(50.79s)$   & $0.00 \%$ & $(1.83m)$ & $0.00\%$\\
 &  Gurobi  &  $0.00\%$ & $(1.24s)$   & $0.00 \%$ & $(30.34s)$  & $0.00 \%$ & $(2.16m)$ & $0.00\%$\\
  & OR-Tools 
 &  $1.96\%$ & $(10.47s)$   & $3.50\%$ & $(1.07m)$ & $4.16 \%$ & $(5.11m)$ & $3.21\%$\\
\cmidrule{2-9}
 &  Farthest Insertion  &  $2.91\%$ & $(1.41s)$   & $6.01\%$ & $(5.73s)$  & $7.49 \%$ & $(9.09s)$ & $5.61\%$\\
 &  Random Insertion  &  $5.10\%$ & $(0.99s)$  & $8.33\%$ & $(3.24s)$  & $10.09 \%$ & $(5.85s)$ & $8.16\%$\\
 & Nearest Insertion  &  $10.94\%$ & $(1.48s)$   & $17.59\%$ & $(5.09s)$  & $20.19 \%$ & $(9.49s)$ & $16.66\%$\\
  \multicolumn{3}{c}{-} \\
 &  LIH(T=1000) 
 &  $14.49\%$ & $(28.24s)$   & $145.03\%$ & $(36.73s)$   & $300.00\%$ & $(2.05m)$ & $176.64\%$ \\
 & GCN(gr.) 
 &  $17.82\%$ & $(17.30s)$   & $75.92\%$ & $(18.55s)$   & $52.92 \%$ & $(1.01m)$ & $48.37\%$\\
   & GCN(bs) 
 &  $16.16\%$ & $(1.03m)$   & $74.36\%$ & $(1.32m)$  & $48.05 \%$ & $(4.97m)$ & $46.70\%$\\
 & MDAM(gr.)  &  $0.72\%$ & $(9.61s)$  & $12.57\%$ & $(32.36s)$   & $13.97 \%$ & $(12.06m)$ & $9.09\%$\\
 & MDAM(bs.)  &  $0.65\%$ & $(52.18s)$   & $5.56\%$ & $(20.95m)$ & $12.99 \%$ & $(38.87m)$ & $5.91\%$\\
 & CVAE-Opt  &  $2.90\%$ & $(1.39m)$   & $7.33\%$ & $(3.01m)$ & $3.62 \%$ & $(4.74m)$ & $4.62\%$\\
 & AM 
 &  $5.95\%$ & $(0.02s)$   & $65.77\%$ & $(0.04s)$  & $53.30 \%$ & $(0.12s)$ & $41.67\%$\\
     & AM(ASP) 
 &  \textbf{$\mathbf{1.81\%}$} & $(0.03s)$  & $\mathbf{3.98\%}$ & $(0.07s)$  & {$\mathbf{5.60\%}$} & $(0.16s)$ & ${\mathbf{3.79\%}}$\\
     & POMO 
 &  \underline{${0.01\%}$} & (0.45s)  & \underline{${0.42\%}$} & (1.43s)  & $2.81\%$ &(5.67s) &$1.09\%$ \\
    & POMO(ASP) 
 &  $0.09\%$ & (0.45s)  & $0.59 \%$ & (1.44s)   & \underline{{$\mathbf{1.67\%}$}} &(5.67s) &\underline{$\mathbf{0.78\%}$} \\
\midrule
\midrule

\multirow{10}{*}{\rotatebox[origin=c]{90}{CVRP}}
 &  LKH3  &  $0.00\%$ & $(4.81m)$  & $0.00\%$ & $(8.23m)$   & $0.00\%$ &$(16.09m)$ &$0.00\%$ \\
 &  Gurobi(10s)  &  $1.74\%$ & $(11.69s)$ & $33.08\%$ & $(12.15s)$   & $40.44\%$ &$(12.70s)$ &$32.38\%$ \\
 &  Gurobi(50s)  &  $1.39\%$ & $(51.80s)$  & $14.93\%$ & $(53.80s)$ & $32.58\%$ &$(53.65s)$ &$14.87\%$ \\
 & Gurobi(100s)  &  $2.78\%$ & $(1.70m)$  & $10.20\%$ & $(1.72m)$  & $21.54\%$ &$(1.78m)$ &$10.82\%$ \\
 
\cmidrule{2-9}
& RL(gr.) 
 &  $42.51\%$ & (2.72s)   & $14.42\%$ & (5.04s)   & $47.94\%$ &(6.12s) &$39.54\%$ \\
 & RL(bs.) 
 &  $34.84\%$ & (6.06s)  & $9.95\%$ & (28.42s)  & $41.57\%$ &(35.48s) &$33.67\%$ \\
  & MDAM(gr.) 
 &  $3.48\%$ & (3.95s)  & $11.95\%$ & (6.15s)  & $12.49\%$ &(18.18s) &$9.31\%$ \\
  & MDAM(bs.) 
 &  $1.74\%$ & (9.48s)  & $9.19\%$ & (24.19s)  & $10.45\%$ &(1.02m) &$7.13\%$ \\
& CVAE-Opt  &  $5.57\%$ & $(4.08m)$   & $9.43\%$ & $(3.71m)$ & $16.79 \%$ & $(3.92m)$ & $10.60\%$\\
 & LoRew 
 &  $29.95\%$ & $(0.34s)$ & $84.70\%$ & $(2.64s)$  &$152.41\%$& $(5.77s)$ &$ 85.08\%$\\
 & AM 
 &  $5.72\%$ & $(0.02s)$   & $11.35\%$ & $(0.06s)$  & $18.09\%$ & $(0.11s)$ & $11.72\%$\\

    & AM(ASP) 
 &  $9.48\%$ & $(0.04s)$  & $\mathbf{7.73\%}$ & $(0.11s)$  & $\mathbf{11.27\%}$ & $(0.21s)$ & $\textbf{9.49\%}$\\
     & POMO 
 &  \underline{$0.32\%$} & (0.04s)  & $1.27\%$ & (0.14s)  & $3.00\%$ &(0.51s) &$1.53\%$ \\

    & POMO(ASP) 
 &  $0.37\%$ & (0.04s)   & \underline{$\mathbf{1.26 \%}$} & (0.11s) & \underline{$\mathbf{2.79\%}$} &(0.44s) &\underline{$\textbf{1.47\%}$} \\
\midrule
\midrule
\multirow{2}{*}{\rotatebox[origin=c]{90}{SDVRP}}

   & MDAM(gr.)  &  $3.83\%$ & $(4.17s)$   & $10.80\%$ & $(9.37s)$ & $16.79 \%$ & $(21.59s)$ & ${10.47\%}$\\
  & MDAM(bs.)  &  \underline{$2.43\%$} & $(11.01s)$ & \underline{$8.27\%$} & $(23.61s)$ & $13.43\%$ & $(1.30m)$ & \underline{${8.03\%}$}\\
     & AM 
 &  $6.20\%$ & (0.03s)   & $26.79\%$ & (0.08s)  & $19.56\%$ &(0.15s) &$17.52\%$ \\
   & AM(ASP) 
 &  $13.32\%$ & (0.05s)  & $\mathbf{8.34\%}$ & (0.13s) & \underline{$\mathbf{11.88\%}$} &(0.26s) &$\mathbf{11.18\%}$ \\
\midrule
\midrule
 
\multirow{6}{*}{\rotatebox[origin=c]{90}{PCTSP}}
 &  Gurobi  &  $0.00\%$ & (0.30s)   & $0.00\%$ & (0.95s)   & $0.00\%$ &(2.98m) &$0.00\%$ \\
      &  OR-Tools &  $2.20\%$ &$(11.06s)$   & $5.33\%$ & (2.00m)   & $8.33\%$ &(6.00m) &$5.30\%$ \\
\cmidrule{2-9}
 & ILS(python 10x) 
 &  $63.23\%$ & (3.05s)   & $148.05\%$ & (4.70s)   & $209.78\%$ &(5.27s) &$137.94\%$ \\
    \multicolumn{3}{c}{-} \\
    & MDAM(gr.) 
 &  $11.76\%$ & $(41.10s)$ & $24.73\%$ & $(1.31m)$   & $30.07\%$ & $(1.96m)$ &$22.19\%$ \\
 & MDAM(bs.) 
 &  $5.88\%$ & $(2.70m)$   & $18.81\%$ & $(4.77m)$   & $26.09\%$ & $(6.97m)$ &$16.93\%$ \\
 
     & AM 
 &  \underline{$2.88\%$} & (0.02s)   & $17.95\%$ & (0.06s)   & $29.24\%$ &(0.14s) &$16.69\%$ \\
    & AM(ASP) 
 &  $12.05\%$ & (0.03s)   & \underline{$\mathbf{10.34 \%}$} & (0.08s)   & \underline{$\mathbf{11.56\%}$} &(0.18s) &\underline{$\mathbf{11.32\%}$} \\

\midrule
\midrule
\multirow{4}{*}{\rotatebox[origin=c]{90}{SPCTSP}}

 & MDAM(gr.) 
 &  $11.76\%$ & $(20.33s)$ & $24.73\%$ & $(11.36s)$   & $30.07\%$ & $(16.97s)$ &$22.19\%$ \\
 & MDAM(bs.) 
 &  $7.35\%$ & $(1.06m)$   & $20.43\%$ & $(2.02m)$   & $26.81\%$ & $(3.10m)$ &$18.20\%$ \\
     & AM 
 &  \underline{$3.60\%$} & (0.02s)  & $15.68\%$ & (0.06s)   & $33.09\%$ &(0.13s) &$17.45\%$ \\
    & AM(ASP) 
 &  $10.74\%$ & (0.04s)   & \underline{$\mathbf{7.69 \%}$} & (0.07s) & \underline{$\mathbf{8.16}\%$} &(0.16s) &\underline{$\mathbf{8.86\%}$} \\
\midrule
\midrule
\end{tabular}
}
\end{table*}

\noindent\textbf{Data Generation} --  
For TSP, We generate data by randomly sampling $x\in\mathbb{R}^2$ from the unit square, and sampling $y\in\mathbb{R}^2$ from $\text{N}(\textbf{0},\Sigma)$ where $\Sigma\in\mathbb{R}^{2\times2}$ is a diagonal matrix whose elements are sampled from $[0,\lambda]$ and $\lambda\sim\text{U}(0,1)$. Next, a two-dimensional coordinate is generated by $z=x+y$, and we can get any scale $n$ of TSP by performing this sampling $n$ times. We sample 10000 normalized instances which make up 10 groups of data generated by different $\lambda$ values. For CVRP, the generation of two-dimension coordinates is same as TSP, the demands $\delta$ is discrete and sampled uniformly from $\{1,...,9\}$ and the capacity is $D^{20}=30,D^{30}=D^{40}=D^{50}=30,D^{60}=D^{70}=...=D^{100}=50$ and the final demand is normalized to $\hat{\delta}=\frac{\delta}{D}$. The evaluation size is 128. 
For PCTSP and SPCTSP, the maximal length is $L^{20}=2,L^{30}=L^{40}=L^{50}=3,L^{60}=L^{70}=...=L^{100}=4$.
Then we keep the same generation setting for the prize and penalty in AM~\cite{kool2018attention}, the generation of two-dimension coordinates is same as TSP and the evaluaiton size is 1000.
The overall results on problem scale of 20, 50, 100 are shown in Table~\ref{tab:results on generated data}. 

\noindent \textbf{Remark} -- It is worth mentioning the comparing scenario at the beginning: we evaluate one \textit{base solver}(ASP) on different problem scales while the other neural solvers are evaluated with models trained on separate problem scales. That is, we use one universal neural solver to achieve better performance than several base solvers (3 in our case).

\noindent\textbf{Results on TSP} -- We compare various baselines with several neural solvers. Firstly, the performance of neural solvers trained on uniform distribution has a huge decrease on unseen distribution. Specifically, in terms of the average gap, the performances of LIH~\cite{lu2019learning}, GCN~\cite{joshi2019efficient}, MDAM \cite{xin2021multi} and AM~\cite{kool2018attention} are worse than the classical heuristics, e.g. Farthest Insertion and Random Insertion. It's especially undesirable as neural solvers need extra cost for training. The performance of CVAE-Opt \cite{hottung2020learning} and POMO~\cite{kwon2020pomo} are relatively stable. For the comparison between the base solvers and our method,
AM(ASP) and POMO(ASP) have a great decrease of $90.9\%$ and $49.4\%$ compared with AM and POMO while the time-consumption is almost no further increase\footnote{The time cost of AM(ASP) is slightly larger than AM because we use more model capacity. The reasons are shown in section \ref{model capacity} and \ref{app:details on dg}}. What's more, POMO(ASP) achieves the best results among all solvers except for the exact methods.

\noindent\textbf{Results on VRP} -- For CVRP, the neural solvers have dominant performance w.r.t. time-consumption compared with the classic solvers. For the solution quality, the neural solvers still suffer from the impacts of data distribution heavily except POMO~\cite{kwon2020pomo}. For the comparison of base solvers and our method, AM(ASP) and POMO(ASP) have a $19\%$ and $5\%$ decrease over AM and POMO without increased time consumption. Furthermore, even compared with Gurobi (10s, 50s, 100s)~\cite{gurobi} and OR-Tools~\cite{ortools} in terms of solution quality and time-consumption, POMO(ASP) achieves the best performance among all these baselines, showing the potential of neural solvers to handle more complex tasks. For SDVRP, AM(ASP) also has a 36.2\% improvement compared with AM.

\noindent\textbf{Results on PCTSP} -- Compared with classical solvers and heuristics, neural solvers still have superiority over the time-consumption. For the solution quality, AM and AM(ASP) outperform Iterated Local Search (ILS) by a large margin. Moreover, AM (ASP)'s performance is better than AM with a $32.2\%$ decrease. Similar to the results on SPCTSP, AM(ASP)'s performance surpasses AM by $29.2\%$, showing a much better generalization ability for real problems.

\subsection{Results on Real World Data}
We show the experimental results on real datasets in Table~\ref{tsp results on real problem}--\ref{cvrp results on real problem}. Note both AM and POMO are trained under the problem scale of 100, requiring far more training time than AM(ASP) and POMO(ASP).

\begin{table}[hbpt!]
\renewcommand{\arraystretch}{.99}
\centering 
\caption{Results on TSPlib instances for TSP. The underlined and bold figures mean achieving better results than OR-Tools and the original base solver, respectively.}
\label{tsp results on real problem}
\scalebox{0.75}{
\begin{tabular}{l|cc|cc|cc} 
\toprule 
\toprule 
\multirow{2}{*}{\rotatebox[origin=c]{0}{\textbf{\makecell{Instance}}}} & \multirow{2}{*}{\rotatebox[origin=c]{0}{\makecell{Opt.}}} &
\multirow{2}{*}{\rotatebox[origin=c]{0}{\makecell{OR-Tools}}} & 
\multirow{2}{*}{\rotatebox[origin=c]{0}{\makecell{AM}}}  &
\multirow{2}{*}{\rotatebox[origin=c]{0}{\makecell{AM(ASP)}}}  &
\multirow{2}{*}{\rotatebox[origin=c]{0}{\makecell{POMO}}} & \multirow{2}{*}{\rotatebox[origin=c]{0}{\makecell{POMO(ASP)}}}  \\ 
 &   &  &  &  &  & \\
\midrule 
pr226 & 80,369  & 82,968  & 86,200  & \textbf{83,570} & \underline{81,903} &83,117\\
ts225 & 126,643  & 128,564  & 143,994  & \textbf{140,783} & 134,184 &138,846\\
kroD100 & 21,294  & 21,636  & 21,913  & \textbf{21,817} & 21,878 &\textbf{\underline{21,508}}\\
eil51 & 426  & 436  & \underline{434}  & 438 &{\underline{428}} & \underline{429}\\
kroA100 & 21,282  & 21,448  & 23,500  & \textbf{21,573} & 21,691 &\textbf{\underline{21,332}}\\
pr264 & 49,135  & 51,954  & 60,952  & \textbf{57,398} & \underline{51,812} &55,643\\
pr152 & 73,682  & 75,834  & 76,984  & 77,008 & \underline{74,088} &\underline{75,086}\\
rat99 & 1,211  & 1,232  & 1,329  & \textbf{1,298} & 1,273 &\textbf{\underline{1,221}}\\
kroA150 & 26,524  & 32,474  & \underline{29,674}  & \underline{\textbf{27,799}} & 27,087 &\underline{\textbf{{26,982}}}\\
lin105 & 14,382  & 14,824  & 16,423  & \underline{\textbf{14,640}} & 14,853 &\underline{\textbf{14,531}}\\
pr124 & 59,030  & 62,519  & \underline{62,350}  & \underline{\textbf{59,981}} & \underline{59,385} &\underline{59,399}\\
st70 & 675  & 683  & 701  & \underline{\textbf{677}} & \underline{677} &\underline{682}\\
a280 & 2,586  & 2,713  & 3,127  & \textbf{2,913} & 2,844 &2,891\\
rd100 & 7,910  & 8,189  & 8,191  & 8,316 & \underline{7,917} &\underline{7,923}\\
pr136 & 96,772  & 102,213  & 103,878  & \underline{\textbf{100,927}} & \underline{98,183} &\underline{99,671}\\
pr76 & 108,159  & 111,104  & \underline{109,958}  & \underline{\textbf{109,123}} & \underline{108,280} &\underline{108,328}\\
kroA200 & 29,368  & 29,714  & 33,033  & \textbf{31,673} & 29,946 &30,313\\
kroB200 & 29,437  & 30,516  & 32,501  & \textbf{31,309} & 30,999 &\textbf{\underline{30,499}}\\
pr107 & 44,303  & 45,072  & 48,115  & \textbf{47,321} & \underline{44,856} &{45,222}\\
kroB150 & 26,130  & 27,572  & 27,837  & \underline{\textbf{26,925}} & 26,680 &\textbf{\underline{26,441}}\\
u159 & 42,080  & 45,778  & \underline{45,302}  & \underline{\textbf{43,178}} & \underline{42,716} &\underline{42,768}\\
berlin52 & 7,542  & 7,945  & 8,483  & \underline{\textbf{7,593}} & \underline{7,548} &\textbf{\underline{7,544}}\\
rat195 & 2,323  & 2,389  & 2,855  & \textbf{2,699} & 2,506 &{2,527}\\
d198 & 15,780  & 15,963  & 24,324  & \textbf{18,482} & 19,864 &\textbf{17,737}\\
eil101 & 642  & 664  & \underline{663}  & \underline{\textbf{659}} & \underline{643} &\underline{643}\\
pr144 & 58,537  & 59,286  & 62,038  & \textbf{59,702} & \underline{59,173} &\underline{59,173}\\
pr299 & 48,191  & 48,447  & 60,506  & \textbf{56,955} & 52,807 &{53,213}\\
kroC100 & 20,750  & 21,583  & 22,491  & \underline{\textbf{20,896}} & 20,906 &\textbf{\underline{20,753}}\\
tsp225 & 3,859  & 4,046  & 4,563  & \textbf{4,361} & 4,146 &4,180\\
eil76 & 538  & 561  & 562  & \underline{\textbf{555}} & \underline{549} &\underline{553}\\
kroB100 & 22,141  & 23,006  & 23,293  & \underline{\textbf{22,923}} & \underline{22,336} &\underline{22,376}\\
kroE100 & 22,068  & 22,598  & 23,311  & \textbf{22,733} & \underline{22,415} &\textbf{\underline{22,183}}\\
ch150 & 6,532  & 6,729  & 6,851  & \textbf{6,756} & \underline{6,590} &\underline{6,609}\\
bier127 & 118,282  & 137,893  & \underline{128,308}  & \underline{\textbf{121,687}} & \underline{124,391} &\underline{\textbf{122,874}}\\
ch130 & 6,110  & 6,284  & 6,452  & \underline{\textbf{6,218}} &\underline{6,137} &\underline{6,137}\\
\midrule 
Avg. Gap ($\%$)  & 0  &4.19 &10.69 & \textbf{5.62} & \underline{3.37}&\textbf{\underline{3.23}}\\ 
\bottomrule 
\bottomrule 
\end{tabular}
}
\end{table}

\begin{table}[hbpt!]
\renewcommand{\arraystretch}{.99}
\caption{Results on CVRPlib Instances for CVRP. The underlined and bold numbers
indicate better results than OR-Tools and the original base solver,
respectively.}
\label{cvrp results on real problem}
\centering 
\scalebox{.75}{
\begin{tabular}{l|cc|cc|cc} 
\toprule 
\toprule 
\multirow{2}{*}{\rotatebox[origin=c]{0}{\textbf{\makecell{Instance}}}} & \multirow{2}{*}{\rotatebox[origin=c]{0}{\makecell{Opt.}}} &
\multirow{2}{*}{\rotatebox[origin=c]{0}{\makecell{OR-Tools}}} & 
\multirow{2}{*}{\rotatebox[origin=c]{0}{\makecell{AM}}}  &
\multirow{2}{*}{\rotatebox[origin=c]{0}{\makecell{AM(ASP)}}}  &
\multirow{2}{*}{\rotatebox[origin=c]{0}{\makecell{POMO}}} & \multirow{2}{*}{\rotatebox[origin=c]{0}{\makecell{POMO(ASP)}}}  \\ 
 &   &  &  &  &  & \\
\midrule 
B-n52-k7 & 747  & 759  & 806  & 850 & \underline{753} &{\underline{755}}\\
B-n66-k9 & 1,316  & 1,390  & 1,437  & \textbf{1,418} & 1,335 &\textbf{\underline{1,331}}\\
B-n68-k9 & 1,272  & 1,389  & \underline{1,316}  & 1,419 & \underline{1,301} &{\underline{1,313}}\\
B-n45-k6 & 678  & 769  & \underline{748}  & \underline{\textbf{735}} & 717 &\textbf{\underline{689}}\\
B-n50-k8 & 1,312  & 1,335  & 1,378  & \textbf{1,358} & \underline{1,329} &1,359\\
B-n57-k9 & 1,598  & 1,689  & \underline{1,674}  & \underline{\textbf{1,647}} & \underline{1,607} &{\underline{1,617}}\\
B-n51-k7 & 1,032  & 1,150  & \underline{1,034}  & 1,165 & \underline{1,030} &\textbf{\underline{1,025}}\\
B-n50-k7 & 741  & 748  & 788  & 857 & 748 &{764}\\
B-n39-k5 & 549  & 590  & \underline{577}  & 645 & \underline{556} &\textbf{\underline{554}}\\
B-n31-k5 & 672  & 677  & 717  & \textbf{702} & 729 &\textbf{681}\\
B-n63-k10 & 1,496  & 1,627  & 1,633  & \underline{\textbf{1,597}} & \underline{1,540} &\underline{1,579}\\
B-n64-k9 & 861  & 1,105  & \underline{1,010}  & \underline{\textbf{915}} & \underline{910} &\underline{924}\\
B-n67-k10 & 1,032  & 1,105  & 1,144  & 1,146 & \underline{1,101} &\textbf{\underline{1,070}}\\
B-n41-k6 & 829  & 861  & 862  & 940 & \underline{843} &\textbf{\underline{842}}\\
B-n44-k7 & 909  & 948  & 1,030  & \textbf{993} & \underline{934} &950\\
B-n56-k7 & 707  & 779  & \underline{742}  & \underline{766} & 715 &\textbf{\underline{721}}\\
B-n34-k5 & 788  & 798  & 825  & 864 & 803 &\textbf{\underline{796}}\\
B-n35-k5 & 955  & 996  & 1,097  & \underline{\textbf{975}} & \underline{976} &\textbf{\underline{974}}\\
B-n45-k5 & 751  & 804  & 864  & \textbf{812} & \underline{763} &\textbf{\underline{762}}\\
B-n38-k6 & 805  & 881  & \underline{847}  & \underline{852} & \underline{820} &\textbf{\underline{817}}\\
B-n78-k10 & 1,221  & 1,299  & 1,336  & \textbf{1,313} & \underline{1,255} &{\underline{1,265}}\\
B-n43-k6 & 742  & 771  & 830  & \underline{\textbf{770}} & 763 &\textbf{\underline{747}}\\
E-n22-k4 & 375  & 375  & 441  & \textbf{408} & 417 &\textbf{\underline{375}}\\
E-n23-k3 & 569  & 570  & 620  & 620 & 702 &\textbf{572}\\
E-n33-k4 & 835  & 937  & \underline{855}  & 956 & \underline{870} &\textbf{\underline{839}}\\
E-n51-k5 & 521  & 587  & \underline{582}  & \underline{\textbf{572}} & \underline{553} &\textbf{\underline{532}}\\
E-n76-k7 & 682  & 740  & \underline{718}  & 764 & \underline{702} &\textbf{\underline{700}}\\
E-n76-k8 & 735  & 824  & \underline{790}  & \underline{822} & \underline{750} &\textbf{\underline{747}}\\
E-n76-k10 & 830  & 992  & \underline{880}  & \underline{926} & \underline{853} &\textbf{\underline{851}}\\
E-n76-k14 & 1021  & 1,586  & \underline{1,110}  & \underline{1,120} & \underline{1,051} &\textbf{\underline{1,042}}\\
E-n101-k8 & 815  & 993  & \underline{863}  & \underline{900} & \underline{845} &{\underline{845}}\\
E-n101-k14 & 1,067  & 1,238  & \underline{1,119}  & 1,242 & \underline{1,118} &\textbf{\underline{1,117}}\\
F-n45-k4 & 724  & 733  & 784  & \textbf{758} & 761 &\textbf{\underline{725}}\\
F-n72-k4 & 237  & 315  & \underline{308}  & \underline{\textbf{267}} & \underline{260} &{\underline{266}}\\
F-n135-k7 & 1,162  & 1,427  & 1,592  & \underline{\textbf{1,398}} & \underline{1,284} &{\underline{1,334}}\\
X-n101-k25 & 27,591  & 29,405  & 30,140  & \textbf{29,783} & 35,824 &\textbf{\underline{29,137}}\\
X-n106-k14 & 26,362  & 27,343  & 27,661  & 28,676 & 27,994 &\textbf{\underline{27,304}}\\
X-n110-k13 & 14,971  & 16,149  & \underline{15,896}  & 16,367 & \underline{15,132} &{\underline{15,366}}\\
X-n115-k10 & 12,747  & 13,320  & 15,155  & \textbf{14,635} & 13,429 &{13,885}\\
X-n120-k6 & 13,332  & 14,242  & 14,709  & 16,167 & \underline{14,137} &\textbf{\underline{13,805}}\\
X-n125-k30 & 55,539  & 58,665  & 61,176  & \textbf{59,603} & 68,326 &\textbf{\underline{58,317}}\\
X-n129-k18 & 28,940  & 31,361  & \underline{30,900}  & \underline{\textbf{30,560}} & \underline{29,450} &{\underline{29,469}}\\
X-n134-k13 & 10,916  & 13,275  & \underline{12,643}  & \underline{\textbf{12,108}} & \underline{11,421} &\textbf{\underline{11,303}}\\
X-n139-k10 & 13,590  & 15,223  & \underline{15,042}  & \underline{\textbf{14,873}} & \underline{13,982} &\textbf{\underline{13,972}}\\
X-n143-k7 & 15,700  & 17,470  & 17,661  & \underline{\textbf{17,450}} & \underline{16,157} &{\underline{16,379}}\\
X-n148-k46 & 43,448  & 46,836  & 52,201  & \textbf{47,545} & 54,698 &\textbf{\underline{45,456}}\\
X-n153-k22 & 21,220  & 22,919  & 25,832  & \textbf{24,449} & 23,581 &\textbf{23,317}\\
X-n157-k13 & 16,876  & 17,309  & 18,923  & \textbf{18,374} & 17,424 &\textbf{17,371}\\
X-n162-k11 & 14,138  & 15,030  & \underline{14,880}  & 15,786 & \underline{15,022} &\textbf{\underline{14,814}}\\
X-n167-k10 & 20,557  & 22,477  & 22,677  & \textbf{22,478} & \underline{21,454} &\textbf{\underline{21,113}}\\
X-n172-k51 & 45,607  & 50,505  & 55,098  & \underline{\textbf{50,044}} & 60,646 &\textbf{\underline{48,850}}\\
X-n176-k26 & 47,812  & 52,111  & 56,449  & 56,582 & 53,150 &\textbf{\underline{52,012}}\\
X-n181-k23 & 25,569  & 26,321  & 27,114  & \textbf{26,951} & 28,176 &\textbf{\underline{26,208}}\\
    \midrule 
Avg. Gap ($\%$)  & 0  &9.52 &10.74  &\textbf{9.88} & \underline{6.10}&\textbf{\underline{3.32}}\\ 
\bottomrule 
\bottomrule 
\end{tabular}
}
\end{table}

\noindent\textbf{Results on TSP} -- We further verify the neural solvers trained by ASP in TSPLib \cite{reinelt1991tsplib}. 
As shown in Table~\ref{tsp results on real problem}, AM(ASP) outperforms AM in almost all instances with an decrease of 5\% on the average optimality gap. Specifically, for several instances: lin105, st70, pr136, kroB150, berlin52, eil101, kroC100, eil76 and ch130, ASP can help AM surpass OR-Tools. For POMO~\cite{kwon2020pomo}, ASP can still improve it to some extent. 
 What's more, POMO(ASP) outperforms all listed methods and has a better average optimality gap. The above results can show that neural solvers trained via ASP can handle difficult and real instances by generalizing to complex distributions and a set of problem scales.

\noindent\textbf{Results on CVRP} -- We compare the solution qualities of the neural solvers trained by ASP with base solvers AM \cite{kool2018attention}, POMO~\cite{kwon2020pomo}, and OR-Tools on the real-world instances from CVRPLib~\cite{uchoa2017new} B, E, F, and X, covering the problem scales range from 22 to 181. 
In these real instances, AM(ASP) has better performance than AM in terms of the average optimality gap. The neural solver POMO(ASP) outperforms POMO~\cite{kwon2020pomo} on approximately $70\%$ (37 out of 53) of the selected CVRPlib instances and achieves a better average gap $3.32\%$ than baseline $6.10\%$. Additionally, POMO(ASP) also achieves better performance than OR-tools on 47 out of 53 instances and has a decrease of $65\%$ in terms of the average gap.

Based on the results on the real benchmark, we can conclude neural solvers trained after ASP can improve the generalization ability observably with the same model capacity and far less training time.

\section{Discussion}
In this section, we conduct various ablation studies about essential training details: training paradigm, task selection strategy, performance threshold, model capacity, and training sample size. Based on these results, we provide a comprehensive view of {the ASP} framework and some training insights for neural solvers for future work. For all the ablation studies, we keep the other settings the same except for the factors we focus on and run them under 5 random generated seeds. 
\subsection{Comparison Between Different Training Paradigm}
\begin{figure}[t]
    \centering
    \includegraphics[width=.48\textwidth]{ 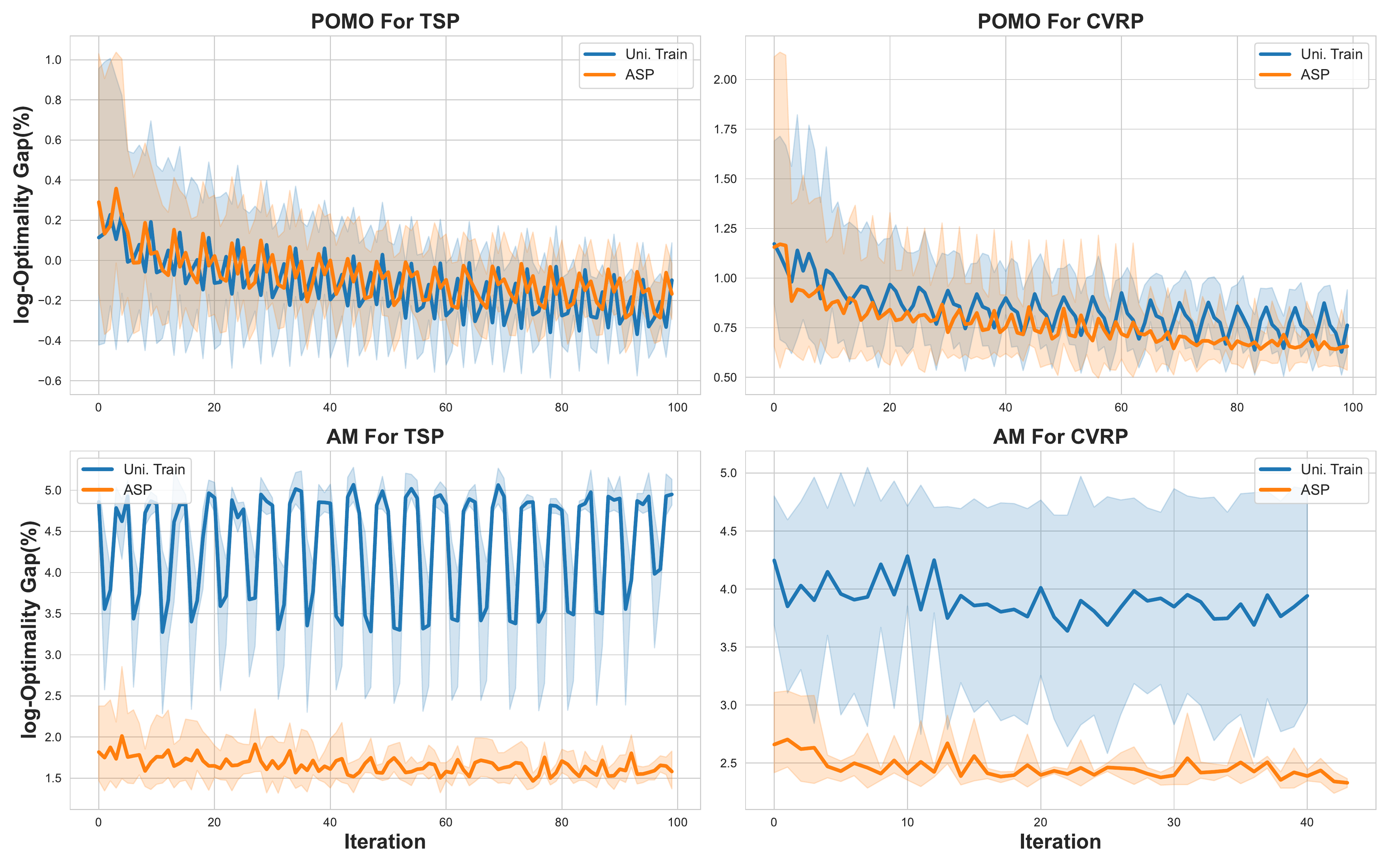}
    \caption{Comparisons of Uniformly Train and ASP. Uniformly train means training all problem scales at the same time and training samples of each problem scale are equal. With Uniformly Train, POMO can achieve competitive results with ASP on TSP and a little worse performance on CVRP. However for AM, the training of Uniformly Train fails on TSP and the performance is poor on CVRP, while ASP can gain consistent improvement for POMO and AM on both TSP and CVRP.}
    \label{fig:uni train vs asp}
\end{figure}
To verify the validity of the ASP framework, we train POMO and AM with mixed problem scales 20, 40, 60, 80, and 100 on uniform distribution, which we denote as ``Uniformly Train''. As demonstrated in Fig. \ref{fig:uni train vs asp}, POMO under uniform training can achieve competitive performance compared with the ASP framework. However, the performances of AM by uniform training are quite worse and the training even fails for TSP and CVRP. This observation indicates that neural solvers do not consistently benefit from a naive training procedure by mixing different types of problem instances. However, the ASP framework shows excellent transferability and thus fits multiple neural solvers and COPs.

\subsection{Influence of Task Selection Strategy}
\begin{figure}[h!]
    \centering
    \includegraphics[width=.48\textwidth]{ 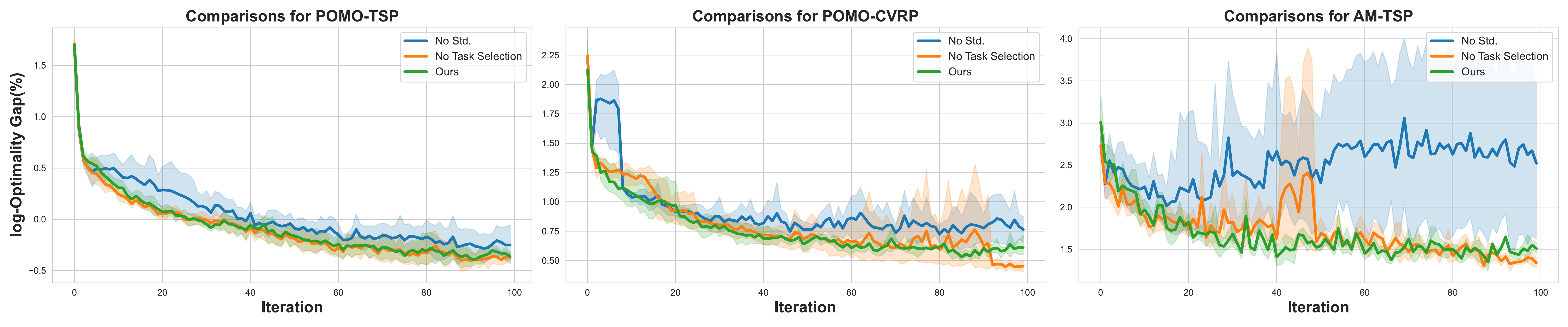}
    \caption{Comparisons of the different training strategies. No Task Selection, No Std. and Ours in the legend mean using uniform strategy, $\lambda=1$ and $\lambda=0.5$ in Eq. \ref{eq: task selection obj} during training. There are cases when the training fails for Task Selection and No Std, e.g. training POMO for CVRP and training AM for TSP. However, ASP shows training stability for all demonstrated cases.}
    \label{fig:sample strategy}
\end{figure}
We propose the task selection strategy in Eq. \ref{eq: task selection obj}:
$$
       p^t_i\propto \lambda p^1_i + (1-\lambda)p^2_i, \lambda\in[0,1]
$$
which receives the information from target distribution to guide the training process. Two naive choices for the task selection strategy are \textbf{(1)} uniform, meaning no preference for specific tasks, and \textbf{(2)} $\lambda=1$, a short-sighted choice to emphasize current performance. Compared with these candidates, our strategy evidently has superiority by taking both the need for improvement and stability into consideration. Experimental results show that these two candidate strategies may fail the training in some cases. As shown in Fig.~\ref{fig:sample strategy}, we evaluate the neural solvers during the training process on the evaluation dataset: Base solver-COP means we train base solvers under the ASP framework on a specific COP with different task selection strategies, "No Std.", "No Task Selection" and "Ours" mean the case of $\lambda=1$, uniform strategy and $\lambda=0.5$, respectively.
Three subfigures show that these three strategies may perform well in some cases, e.g. in the case of POMO-TSP in the leftmost subfigure. However, "No Std" and "No Tasks Selection" show the instability under different random seeds when training POMO-CVRP and even fail the training for AM-TSP because the shortsighted strategy of only considering improvement and uniform strategy can mislead the training. However, ASP performs stably and works well for all neural solvers and COPs under different random seeds.

\subsection{Influence of Performance Threshold}
\begin{figure}[h]
    \centering
    \includegraphics[width=.48\textwidth]{ 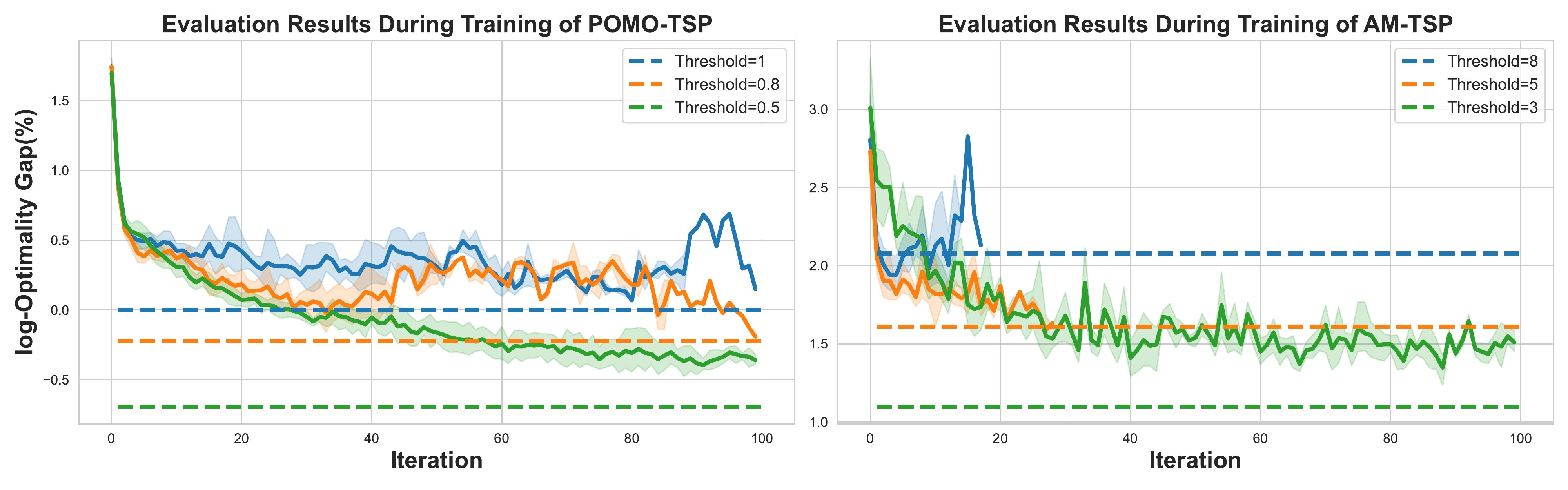}
    \caption{Comparisons of different thresholds for POMO-TSP and AM-TSP w.r.t. the optimality gap (\%) in the log-scale. A lower standard (larger threshold here) will lead to the early stopping of training, for example, AM-TSP achieves the threshold of 8 at a very early iteration. What's more, proper setting of the threshold can improve the performance, for example, POMO-TSP can gain better performance with a threshold of 0.5 than that of 0.8, even though the threshold of 0.5 is not reached eventually.}
    \label{fig: threshold}
\end{figure}

The appropriate threshold is beneficial to the performance a lot. Excessive expectations can lead to indecision while low expectations would cause slacking off. Fig.~\ref{fig: threshold} shows the evaluation curves of neural solvers during the training phase under different performance thresholds. For POMO-TSP, we select three performance thresholds: 1, 0.8, 0.5 and keep the other settings the same; for AM-TSP, we select threshold=8, 5, 3 for comparisons. On one hand, if the performance threshold is too large, the training process of {the ASP} framework will stop in advance, resulting in fewer iterations of the PSA procedure (Algorithm \ref{alg:task selection}). For example, when we set threshold=8 for AM-TSP, the training process stops at very early iteration. On the other hand, if the threshold is too harsh, the neural solver would never achieve the expectation due to its limits, e.g. model capacity, training sample size, etc. For instance, threshold=0.5 for POMO-TSP and threshold=3 for AM-TSP are so rigorous that they can't achieve their goal at the end of training. However, harsh thresholds still promote the neural solver to gain better performance. For instance, even though POMO-TSP is not able to achieve the threshold of 0.5, it can still have better performance on that of training with threshold=0.8.

An obvious guidance of designing the threshold is the performance of the original neural solver which is fixed throughout the whole training process. We believe that a dynamic threshold adjustment procedure will help a lot because the neural solver has different performances on different problem scales.

\subsection{Influence of Model Capacity}\label{model capacity}
\begin{figure}[h]
  \begin{center}
        \includegraphics[width=.48\textwidth]{ 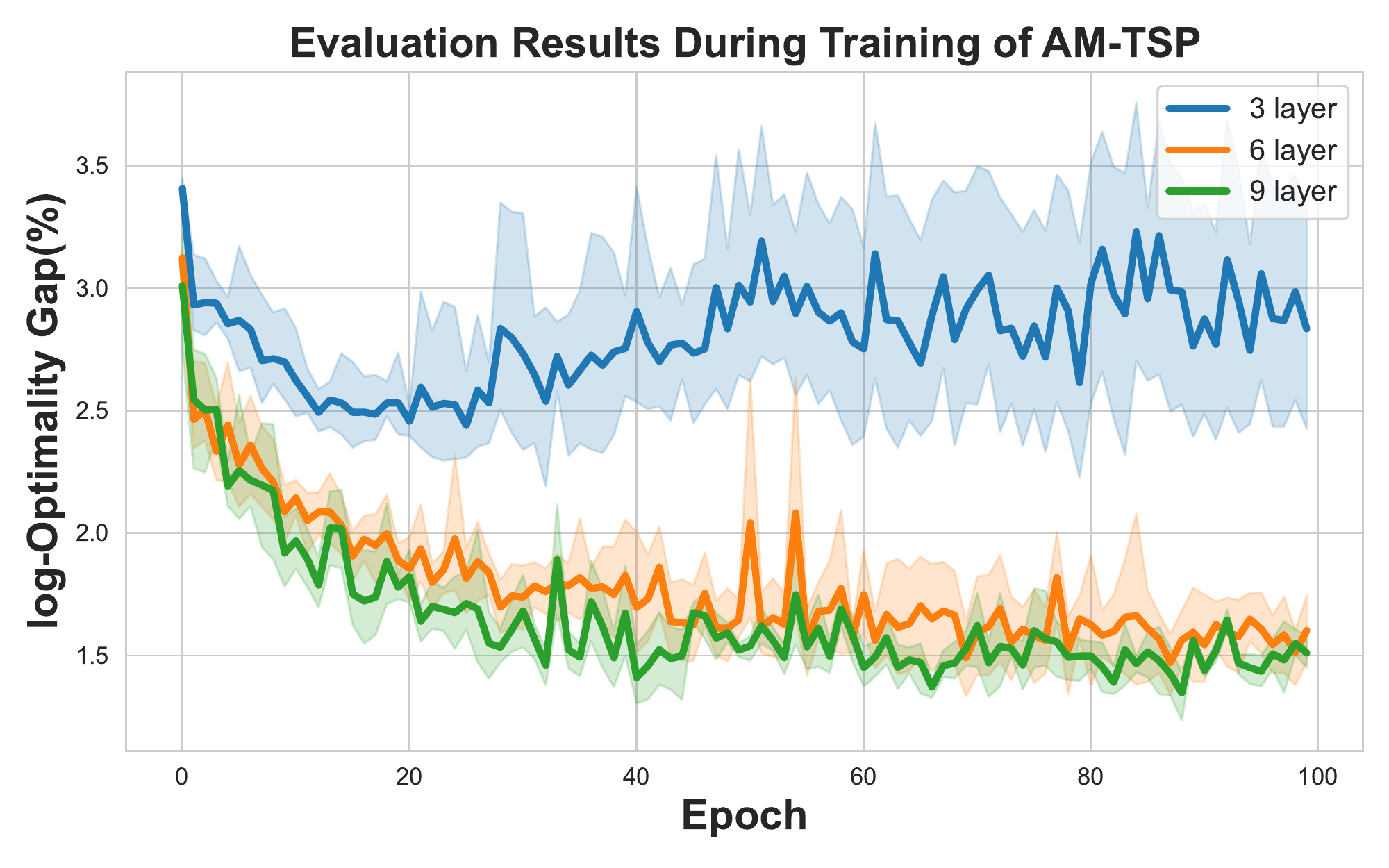}
  \end{center}
  \caption{Evaluation results of AM-TSP with different encoder layers w.r.t. optimality gap (\%) in the log-scale. We can see that a small model capacity will lead to worse performance and even the failure of training.}
  \label{fig:model capacity}
\end{figure}

During the whole training process, we are about to handle a set of problem scales and more complicated distributions which is much harder than solely focusing on a specific problem scale and uniform distribution. When dealing with harder tasks, model capacity is a significant facet impacting the final results, as shown in Fig.~\ref{fig:model capacity}. We can see that AM with 3 encoder layers fails the training while AM with 6 and 9 layers can converge stably, which in turn verifies our assumption. What's more, AM-TSP with 9 encoder layers can obtain better results than that with 6 layers, meaning better performance comes with a larger capacity.

\subsection{Influence of Training Sample Size}
\begin{figure}[t]
    \centering
    \includegraphics[width=.48\textwidth]{ 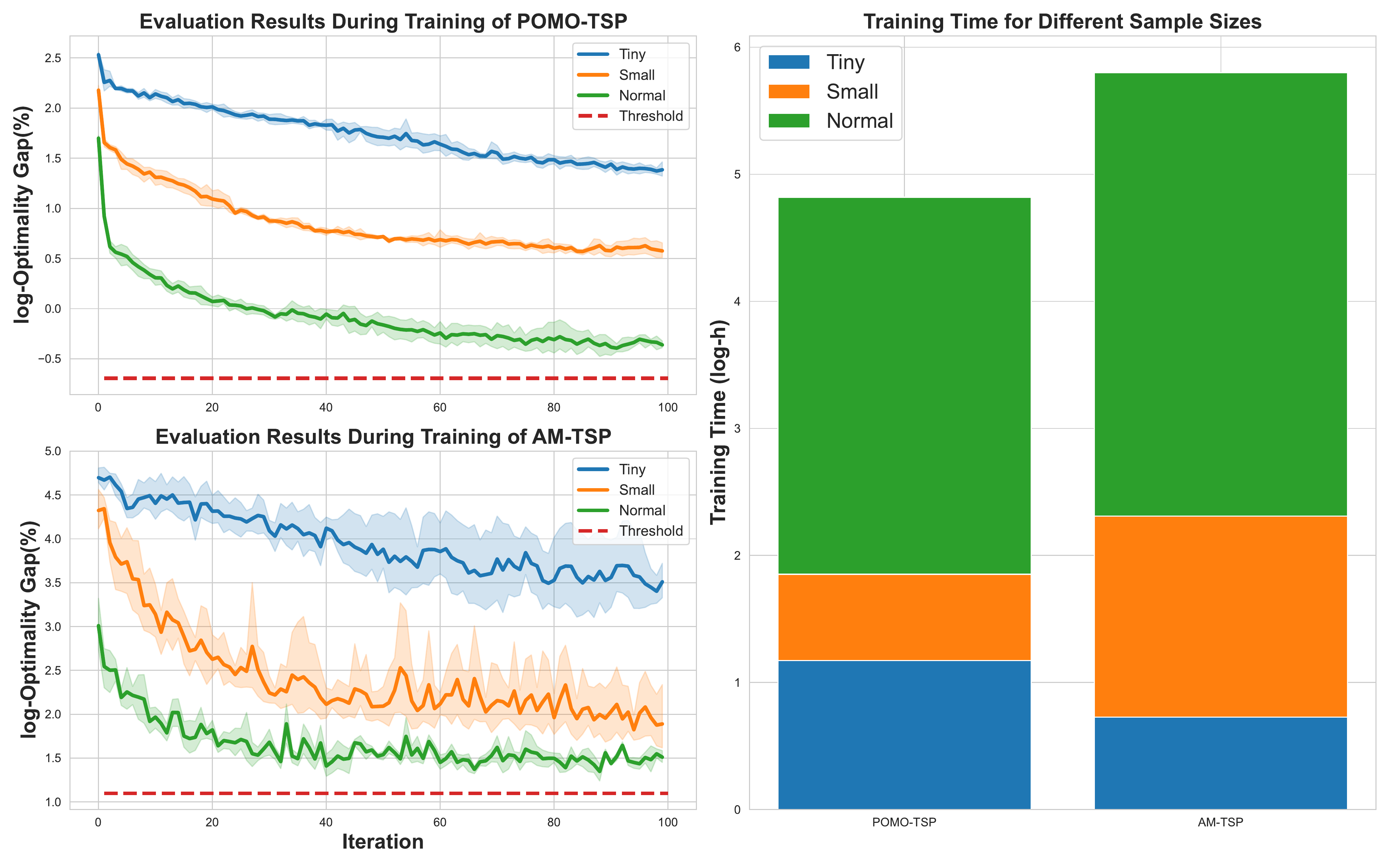}
    \caption{Comparisons of the different sample size. Normal, Small, and Tiny mean the different scales of sample size. The Normal sample size is 10 times larger than Small and The small sample size is 10 times larger than Tiny. The left subfigures show the evaluation results during training and the right one shows the average training time of different random seeds in the log-scale of hours.}
    \label{fig:sample size}
\end{figure}

The training sample size is also an important factor influencing the final performance. We show the ablation results for POMO-TSP and AM-TSP in Fig.~\ref{fig:sample size} and the performance threshold is 0.5 for POMO-TSP and 3 for AM-TSP. In the legend of this figure, ``Normal'' means the sample size $N$ in the default setting of our training, ``Small'' and ``Tiny'' mean the sample size of $\frac{N}{10}$ and $\frac{N}{100}$, respectively. The same threshold can act as a standard to make the comparison clear and fair. It's obvious that a larger sample size remarkably contributes to the final results on the convergence rate and stopping iterations. However, a larger sample size also implies more computational burden. The right bar plot in Fig. \ref{fig:sample size} shows the average training time consumption for different sample sizes in the log-hour scale, so we need to balance this trade-off in realistic cases.

\subsection{Effects of ASP on the Landscape}

There has been a widely-discussed claim whether the shape of the metric landscape is related to the generalization ability \cite{dinh2017sharp, kawaguchi2017generalization,  keskar2016large, chen2020stabilizing}. Some researchers think that ``sharp'' landscape would have poor generalization \cite{chaudhari2019entropy, hochreiter1997flat} due to the large batch size during training, but some others propose specialized training methods to achieve good generalization ability with large batch size \cite{hochreiter1997flat}. In this part, we dive into the generalization ability of AM and POMO by visualizing the return landscape and discussing some insights. We leave specific visualization configurations and further visualization results in Appendix \ref{app: landscape}.

\begin{figure}[ht]
    \centering
    \includegraphics[width=.49\textwidth]{  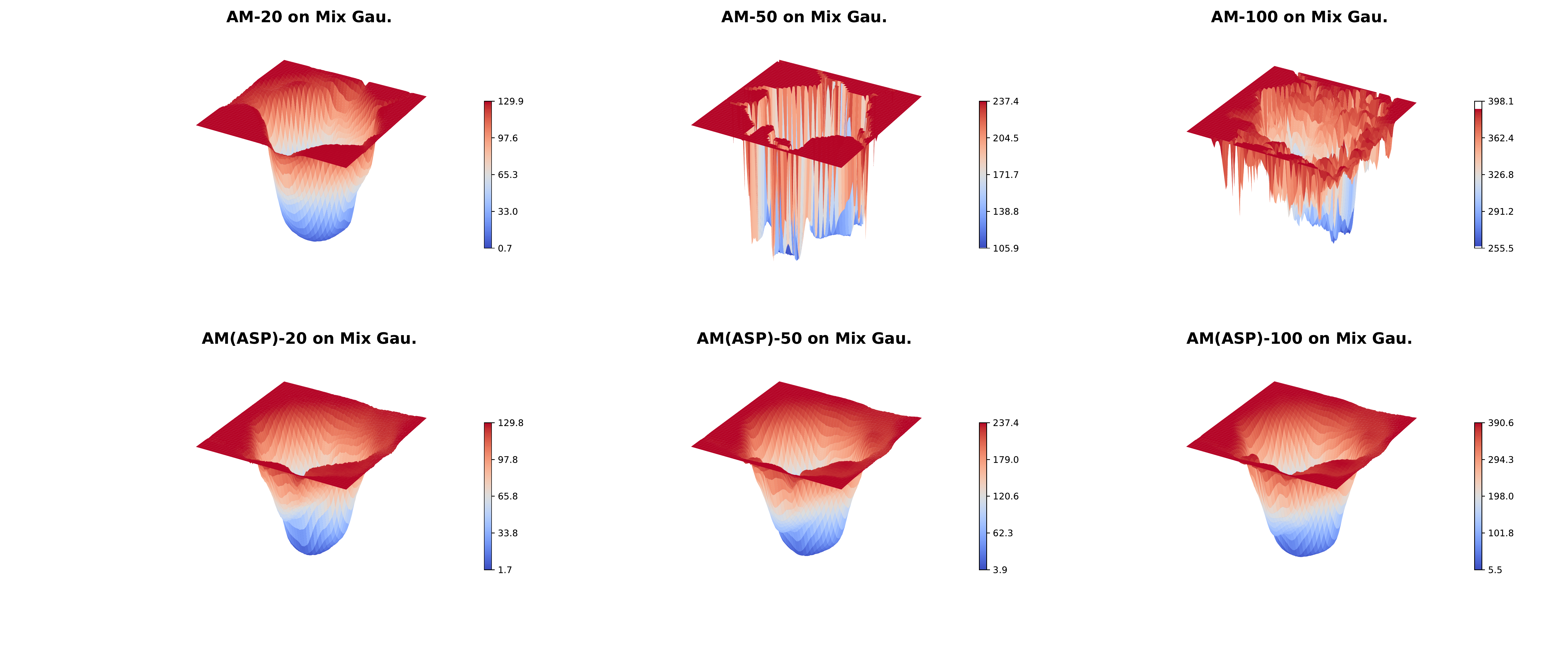}
    \caption{The landscape of optimality gap($\%$) for AM testing on mixed gaussian distribution.}
    \label{fig: am landscape part}
\end{figure}

\noindent\textbf{Discussion for AM.} We visualize the landscape of optimality gap($\%$) for AM and AM(ASP) on the mixed gaussian distribution in Fig. \ref{fig: am landscape part}. Under the same perturbation settings on the parameter, we can see that as the problem scale grows, the landscape is more steep or sharp. For example, the ranges of optimality gap($\%$) for AM(ASP) on TSP20, TSP50, and TSP100 are $1.7\%\sim 129.8\%$, $3.9\%\sim 237.4\%$ and $5.5\%\sim 390.6\%$, respectively, showing that the bigger the problem scale is, the wider the range becomes. What's more, sharper minima means it's more difficult to achieve \cite{li2018visualizing}, so we need to pay more attention to the optimization level or spend more time on training. 
Combining the prior knowledge that larger-scale COPs are harder than smaller ones, we can conclude that large-scale COPs (harder tasks) are more difficult to train than small-scale COPs (easier tasks), which is accordant with the common practice: we always need more training time and get worse performance on large-scale COPs than those on small-scale COPs. Interestingly, the optimization level's observation also coincides with the rule of starting from easy tasks in curriculum training: what curriculum training does is to start from a ``wide minima'' to a ``sharp minima''.

\noindent As for the comparison between AM and AM(ASP), we can see that the landscapes of AM on TSP50 and TSP100 are quite messy, while that on TSP20 is relatively smooth, which aligns with the results in Table \ref{tab:results on generated data} and \ref{tab:further results of AM} that AM performs well on TSP20 but poorly on TSP50 and TSP100. However, all the landscapes obtained from AM(ASP) share the smoothness for all problem scales.

\begin{figure}[t]
    \centering
    \includegraphics[width=.49\textwidth]{  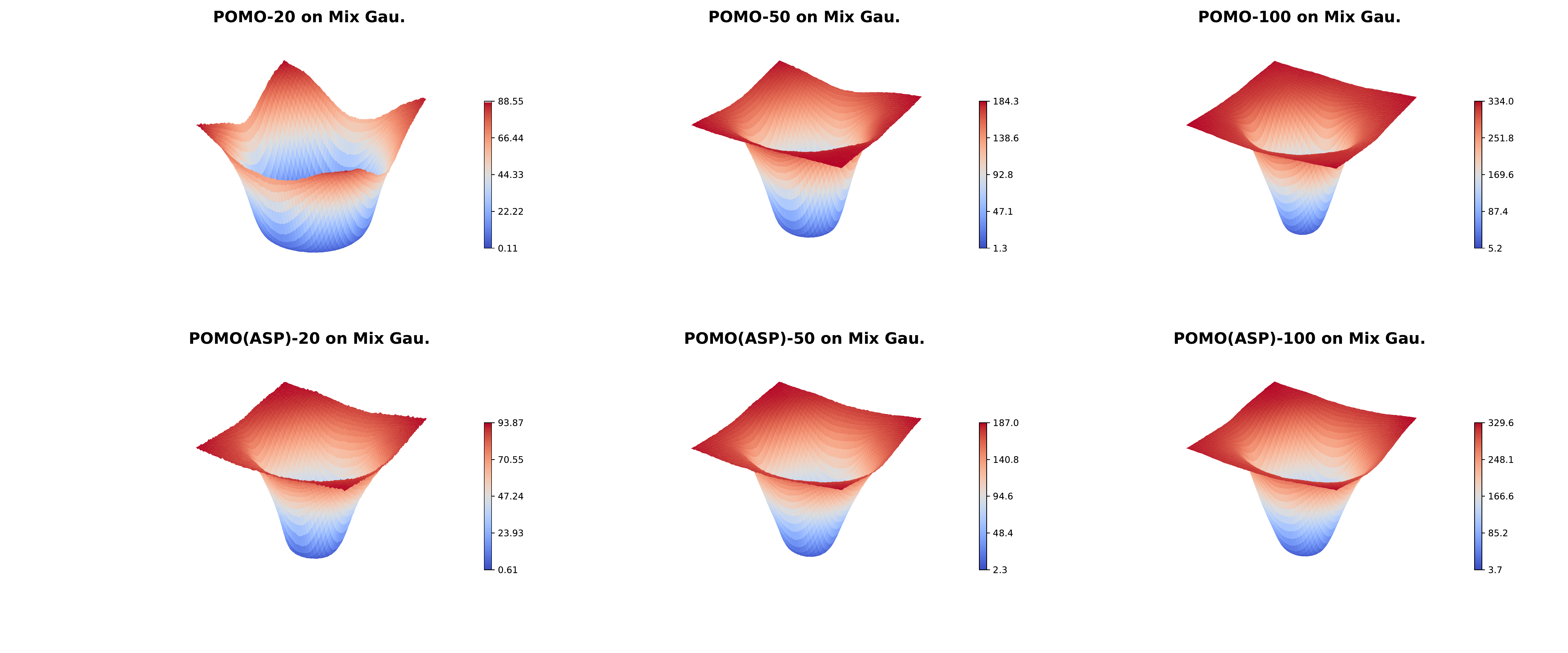}
    \caption{The landscape of optimality gap($\%$) for POMO testing on mixed gaussian distribution.}
    \label{fig: pomo landscape part}
\end{figure}
\noindent\textbf{Discussion for POMO.} We visualize the landscape of optimality gap($\%$) for POMO and POMO(ASP) on the mixed gaussian distribution in Fig. \ref{fig: pomo landscape part}. The relationship between the generalization ability on the problem scale and the curvature of the landscape for POMO shares the same rule: a bigger problem scale has a sharper landscape of optimality gap. Moreover, unlike AM, POMO has a smooth landscape on all problem scales like POMO(ASP), which is accordant to the results in Table \ref{tab:results on generated data} and \ref{tab:further results of AM}. According to the shape of the landscape, we may say that POMO is a more powerful neural solver than AM in some sense.

\section{Conclusion}
Inspired by the framework of PSRO and Adaptive Staircase, in this paper, we propose the game-theoretic and curriculum learning-based solution, ASP, which for the first time improves the generalization ability from multiple aspects for the RL-based neural solvers on COPs.
We evaluate our method on several typical COPs and employ different neural solvers.
On both randomly-generated and real-world benchmarks, we show that the neural solver trained under our framework demonstrates impressive generalization performance when compared to a series of baselines.  
As such, a neural solver trained via ASP may avoid over-fitting and consistently achieve robust performance even on the distribution and problem scales it has never trained, which has never been accomplished in existing neural solvers.
Therefore, we rationally believe that ASP can lead to the first step in applying neural solvers to industrial use.


%
\bibliographystyle{ieeetr}
\bibliography{ref}

\ifCLASSOPTIONcompsoc
  \section*{Acknowledgments}
\else
  \section*{Acknowledgment}
\fi

This work is supported in part by the National Science and Technology Innovation 2030 -- Major program of ``New generation of artificial intelligence'' (No. 2022ZD0116408).

\newpage

\appendices
\section{Oracle Training}\label{app:derive gradient}

In the algorithm~\ref{alg:PSRO}, we need to train two oracles: $S^{'}$ and $\mathbf{P}^{'}_{\mathcal{I}^n}$ as a new policy to be added to the corresponding policy set. Here we will provide a specific derivation about the gradient for training the oracle.

Taken the formula from Eq.~\ref{gradient of LSS}, the gradient is apparent to get:
\begin{equation}\label{derive LSS}
\begin{aligned}
    &\nabla_{\theta}L_{\text{SS}}(\theta)
    =\nabla_{\theta}\mathbf{E}_{\mathbf{P}_{\mathcal{I}^n}\sim\sigma_{\text{DG}}}\mathbf{E}_{{\mathcal{I}^n}\sim\mathbf{P}_{\mathcal{I}^n}}g(S_\theta,{\mathcal{I}^n},\text{Oracle})\\
    &=\mathbf{E}_{\mathbf{P}_{\mathcal{I}^n}\sim\sigma_{\text{DG}}}\mathbf{E}_{{\mathcal{I}^n}\sim\mathbf{P}_{\mathcal{I}^n}}\nabla_{\theta}g(S_\theta,{\mathcal{I}^n},\text{Oracle})\\
    &=\mathbf{E}_{\mathbf{P}_{\mathcal{I}^n}\sim\sigma_{\text{DG}}}\mathbf{E}_{{\mathcal{I}^n}\sim\mathbf{P}_{\mathcal{I}^n}}\frac{\nabla_{\theta}S_\theta(\mathcal{I}^n)}{\text{Oracle}(\mathcal{I})}.
\end{aligned}
\end{equation}
Also for Eq.~\ref{gradient of attack1 TSP}, the computation of this gradient is:
\begin{equation}\label{derive LDG}
\begin{aligned}
    &\nabla_{\gamma}L_{\text{DG}}(\gamma)=\mathbf{E}_{S\sim\sigma_{\text{SS}}}\nabla_{\gamma}\mathbf{E}_{{\mathcal{I}^n}\sim\mathbf{P}_{\mathcal{I}^n,\gamma}}g(S,{\mathcal{I}^n},\text{Oracle})\\
    &=\mathbf{E}_{S\sim\sigma_{\text{SS}}}\int_{\mathcal{I}}\nabla_{\gamma}\mathbf{P}_{\mathcal{I}^n,\gamma}(\mathcal{I})g(S,{\mathcal{I}},\text{Oracle})\mathbf{d}\mathcal{I}\\
    &=\mathbf{E}_{S\sim\sigma_{\text{SS}}}\int_{\mathcal{I}}\mathbf{P}_{\mathcal{I}^n,\gamma}(\mathcal{I})\frac{\nabla_{\gamma}\mathbf{P}_{\mathcal{I}^n,\gamma}(\mathcal{I})}{\mathbf{P}_{\mathcal{I}^n,\gamma}(\mathcal{I})}
    g(S,{\mathcal{I}},\text{Oracle})\mathbf{d}\mathcal{I}\\
    &=\mathbf{E}_{S\sim\sigma_{\text{SS}}}\mathbf{E}_{{\mathcal{I}}\sim\mathbf{P}_{\mathcal{I},\gamma}}\nabla_{\gamma}\log\mathbf{P}_{\mathcal{I}^n,\gamma}(\mathcal{I})g(S,{\mathcal{I}},\text{Oracle}).
\end{aligned}
\end{equation}

\section{Training Details}\label{app:details on dg}
\subsection{Training Setting of Neural Solvers}
\begin{table}[h!]
\centering
\begin{tabular}{lllll}
\toprule
\multicolumn{1}{c}{\bf Base Solver}   &\multicolumn{1}{c}{\bf Encoder layer} &\multicolumn{1}{c}{\bf Batch Size} &\multicolumn{1}{c}{\bf Sample Size} &\multicolumn{1}{c}{\bf Epochs}
\\ 
\midrule
AM         &9 &512 &512$\times$2500 &600\\                                       

POMO      &6 &64 &100$\times$1000 &600\\                                       

\bottomrule
\end{tabular}
\caption{Training configuration for AM \cite{kool2018attention} and POMO \cite{kwon2020pomo} on all COP.}
\label{training setting for neural solvers}
\end{table}

For training neural solvers, the ASP framework runs the DE procedure (algorithm \ref{alg:PSRO}) or PSA procedure (algorithm \ref{alg:task selection}) at each iteration, we run 5 iterations of the DE procedure and 1 iteration of PSA procedure, each iteration contains 5 training epochs. We train 100 iterations so the overall training epochs for the neural solver is 600. To make a fair comparison, in Table \ref{tab:results on generated data}, \ref{tsp results on real problem} and \ref{cvrp results on real problem}, the results of POMO are obtained from the same training setting with POMO(ASP). For AM, we use the original training setting (encoder layer is equal to 3 and the number of training epochs is 100) and the pre-trained model parameters provided by the author because we find the performance of AM trained under the same setting (9 encoder layers and 600 epochs) with AM(ASP) is quite messy. We show the results in Appendix \ref{app: am 600 epochs}.
The specific setting of some key parameters is shown in Table \ref{training setting for neural solvers}.

\subsection{Training Setting of Data Generator}
We use Real-NVP~\cite{dinh2016density} $G_{\gamma}$, a typical Normalizing Flow-based generative model to generate data distribution. Different COPs need the specification of the data generation process.

\noindent\textbf{For TSP} - In classical euclidean TSP, we only need to generate two-dimension coordinates. Specifically, we sample $x\sim\mathbf{U}([0,1]\times[0,1])$ and get data $y=G_{\gamma}(x)\sim P_{g,\gamma}$.

\noindent\textbf{For VRP} - Compared with TSP, the CVRP and SDVRP contain extra constraints on customer demand and vehicle capacity besides the two-dimension coordinates. In the general setting of applying neural solvers to solve CVRP and SDVRP, it's common to set these constraints in a relatively safe manner. To avoid infeasible or bad instances due to the two constraints, we keep the generation of customer demand and vehicles the same as previous works \cite{kool2018attention} and focus on the impacts of euclidean coordinates. Same as TSP, we sample the customer and depot coordinates $x$ from a two-dimension unit square and get data $y=G_{\gamma}(x)\sim P_{g,\gamma}$.

\noindent\textbf{For PCTSP and SPCTSP} -  We keep the same setting in AM~\cite{kool2018attention} for the generation of prize and penalty and focus on the two-dimension coordinates as in TSP, VRP, and OP.

\section{Further Results of AM}\label{app: am 600 epochs}
\begin{table}[h]
\caption{Further results of AM with 9 encoder layers after training 600 epochs. The gap \% is w.r.t. the best value across all methods.}
\label{tab:further results of AM}
\centering
\footnotesize
\setlength{\tabcolsep}{0.35em}
\renewcommand{\arraystretch}{0.8}
\begin{tabular}{ll|rrrrr}
\toprule
 &  
 & \multicolumn{1}{c}{$n=20$} & \multicolumn{1}{c}{$n=40$}  
 & \multicolumn{1}{c}{$n=60$} & \multicolumn{1}{c}{$n=80$} 
 &  \multicolumn{1}{c}{$n=100$}\\

\midrule
\multirow{3}{*}{\rotatebox[origin=c]{90}{\tiny TSP}}
&AM-20 &  $0.73\%$ & $2.09\%$   & $8.02\%$ & $12.33\%$   & $18.09\%$ \\
&AM-50 &  $83.35\%$ & $115.83\%$   & $150.54\%$ & $198.55\%$   & $235.07\%$ \\
&AM-100 &  $117.38\%$ & $165.39\%$   & $200.92\%$ & $281.61\%$   & $312.47\%$ \\
\midrule

  \multirow{3}{*}{\rotatebox[origin=c]{90}{\tiny CVRP}}
&AM-20 &  $3.64\%$ & $10.27\%$   & $19.01\%$ & $33.31\%$   & $31.38\%$ \\
&AM-50 &  $223.24\%$ & $277.00\%$   & $451.86\%$ & $407.61\%$   & $414.29\%$ \\
&AM-100 &  $223.24\%$ & $277.00\%$   & $451.86\%$ & $407.61\%$   & $414.29\%$ \\
\midrule
  \multirow{3}{*}{\rotatebox[origin=c]{90}{\tiny SDVRP}}
&AM-20 &  $3.35\%$ & $10.85\%$   & $17.20\%$ & $27.77\%$   & $28.38\%$ \\
&AM-50 &  $223.24\%$ & $277.00\%$   & $451.86\%$ & $407.61\%$   & $414.29\%$ \\
&AM-100 &  $148.69\%$ & $211.00\%$   & $254.66\%$ & $267.96\%$   & $297.00\%$ \\
\midrule

  \multirow{3}{*}{\rotatebox[origin=c]{90}{\tiny PCTSP}}
&AM-20 &  $0.26\%$ & $22.07\%$   & $53.28\%$ & $66.86\%$   & $81.58\%$ \\
&AM-50 &  $38.12\%$ & $15.11\%$   & $13.76\%$ & $7.14\%$   & $11.77\%$ \\
&AM-100 &  $148.69\%$ & $211.00\%$   & $254.66\%$ & $267.96\%$   & $297.00\%$ \\
\midrule

  \multirow{3}{*}{\rotatebox[origin=c]{90}{\tiny SPCTSP}}
&AM-20 &  $1.29\%$ & $21.77\%$   & $46.33\%$ & $63.09\%$   & $68.25\%$ \\
&AM-50 &  $38.12\%$ & $15.11\%$   & $13.76\%$ & $7.14\%$   & $11.77\%$ \\
&AM-100 &  $148.69\%$ & $211.00\%$   & $254.66\%$ & $267.96\%$   & $297.00\%$ \\
\bottomrule
\end{tabular}
\end{table}

In Table \ref{tab:further results of AM}, we show the results of AM \cite{kool2018attention} trained from 9 encoder layers and 600 epochs on uniform distribution. AM-N means training AM under problem scale N and we evaluate AM-N on the generated instances with problem scales 20, 40, 60, 80, and 100. Results show that AM-50 and AM-100 work poorly due to the overfitting caused by the large training epochs. However, it's interesting that AM-20 seems less affected by this problem.

\section{Generalization Demonstration}
\begin{figure*}[hbpt!]
    \centering
    \includegraphics[width=\textwidth]{  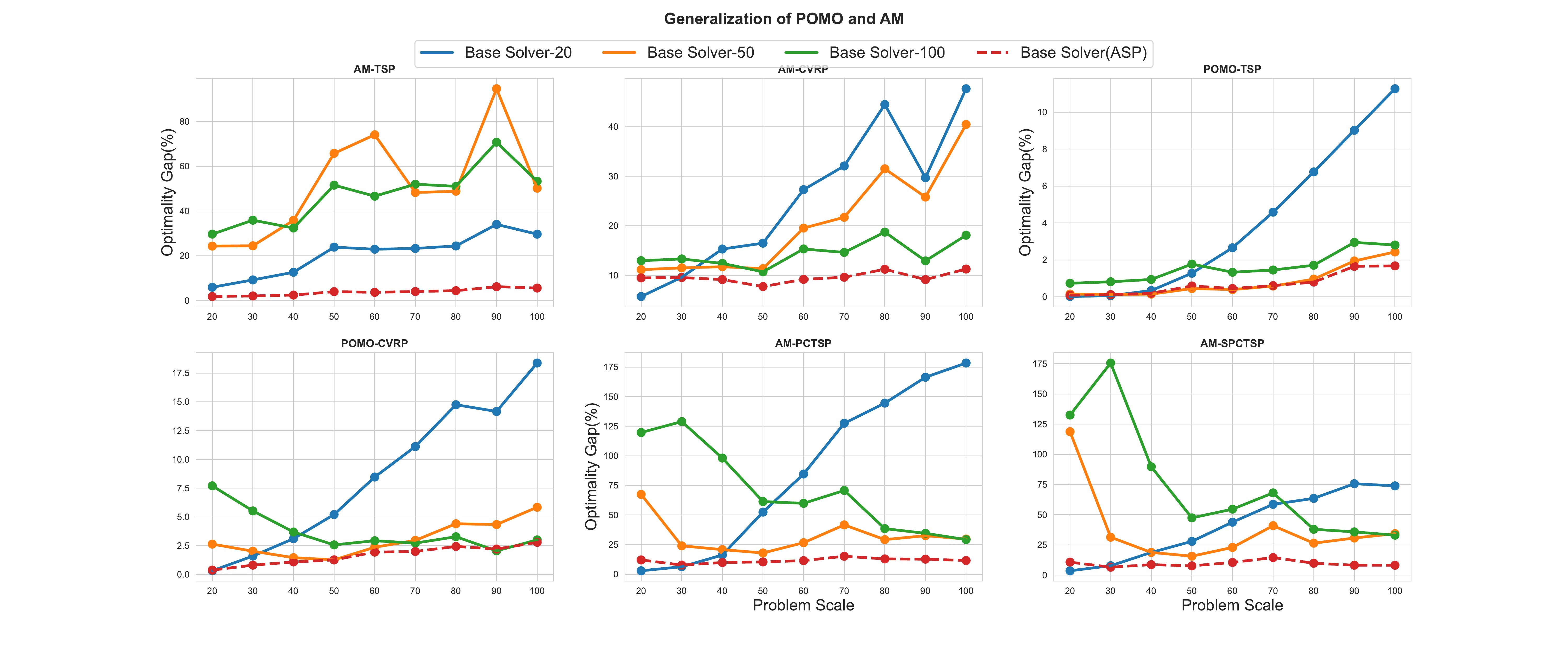}
    \caption{Demonstration of generalization ability. Optimality gap(\%) of AM and POMO as a function of problem scales $n\in\{20,30,40,50,60,70,80,90,100\}$. Results of Base Solver(ASP) are drawn using dashed lines while Base Solvers are drawn with a solid line. Intuitively, the dash (results of Base Solver(ASP)) line are approximately under all the solid lines (results of Base Solver).}
    \label{fig:generalization}
\end{figure*}
We show the generalization results of AM \cite{kool2018attention} and POMO \cite{kwon2020pomo} on different problem scales 20,30,40,50,60,70,80,90 and 100 in Fig. \ref{fig:generalization}. On different problem scales, the base solver's performance degrades as the difference becomes bigger, however, the base solver(ASP) can cover all the problem scales well.

\section{Visualization of Weakness Distribution For TSP}
\begin{figure*}[h]
     \centering
               \begin{subfigure}[b]{\textwidth}
         \centering
         \includegraphics[width=\textwidth]{ 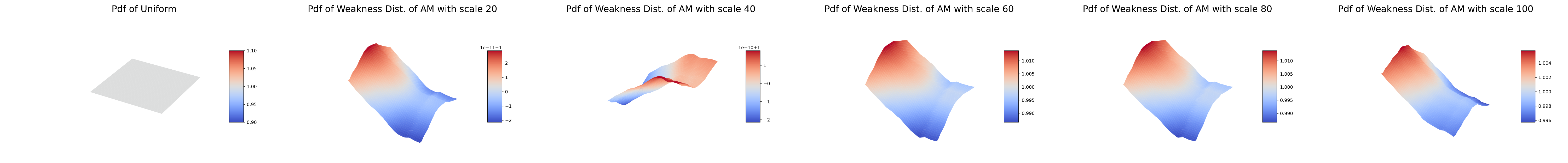}
     \end{subfigure}
     \hfill
     \begin{subfigure}[b]{\textwidth}
         \centering
         \includegraphics[width=\textwidth]{ 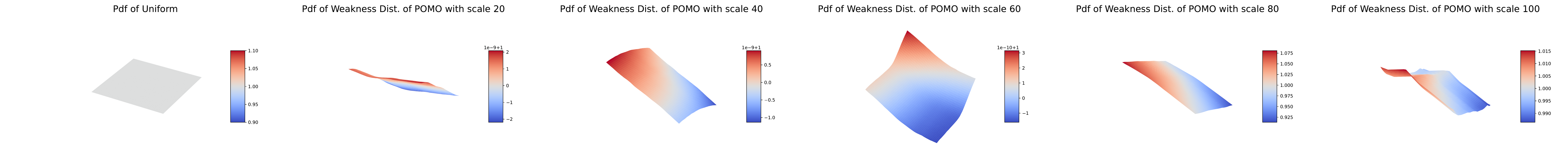}
     \end{subfigure}
    \caption{Demonstration of weakness distribution for AM and POMO. The probability density function (pdf) is taken within the range of $[0,1]\times [0,1]$.}
    \label{fig:weak dist}
\end{figure*}

In this part, we demonstrate the weakness distribution found by our data generator for TSP. From algorithm \ref{alg:PSRO}, the weakness distribution is 
$$\mathbf{P^n}=\sum_i \sigma_{\text{DG},i}\mathbf{P}_{\mathcal{I}^n,i}$$
where $\mathbf{P}_{\mathcal{I}^n,i}$ is the oracle of data generator trained at each PSRO iteration and $\sigma_{\text{DG},i}$ is the meta-strategy. Thanks to the usage of Normalizing Flows, we can compute the probability exactly.
The probability density function (pdf) is obtained within the range of $[0,1]\times[0,1]$. In Fig.~\ref{fig:weak dist}, we can see that the weakness distributions for AM and POMO have an infinitesimal change over the uniform distribution, which can be seen as a specific type of data augmentation.

\section{Further Demonstration of Landscape}\label{app: landscape}
\begin{figure}[hbpt!]
    \centering
    \includegraphics[width=.5\textwidth]{  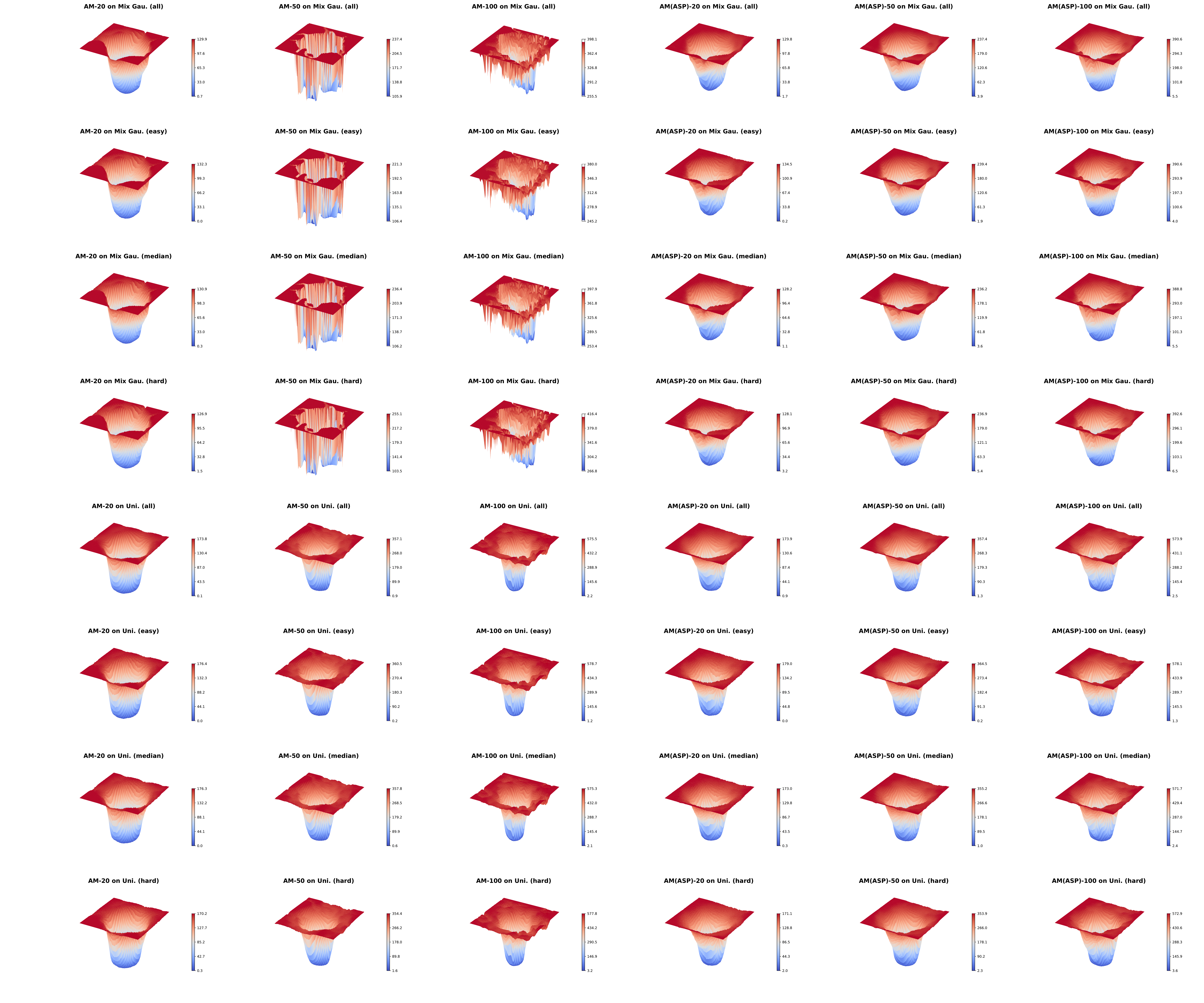}
    \caption{All the landscapes of optimality gap (\%) for AM.}
    \label{fig:am landscape}
\end{figure}

\begin{figure}[hbpt!]
    \centering
    \includegraphics[width=.5\textwidth]{ 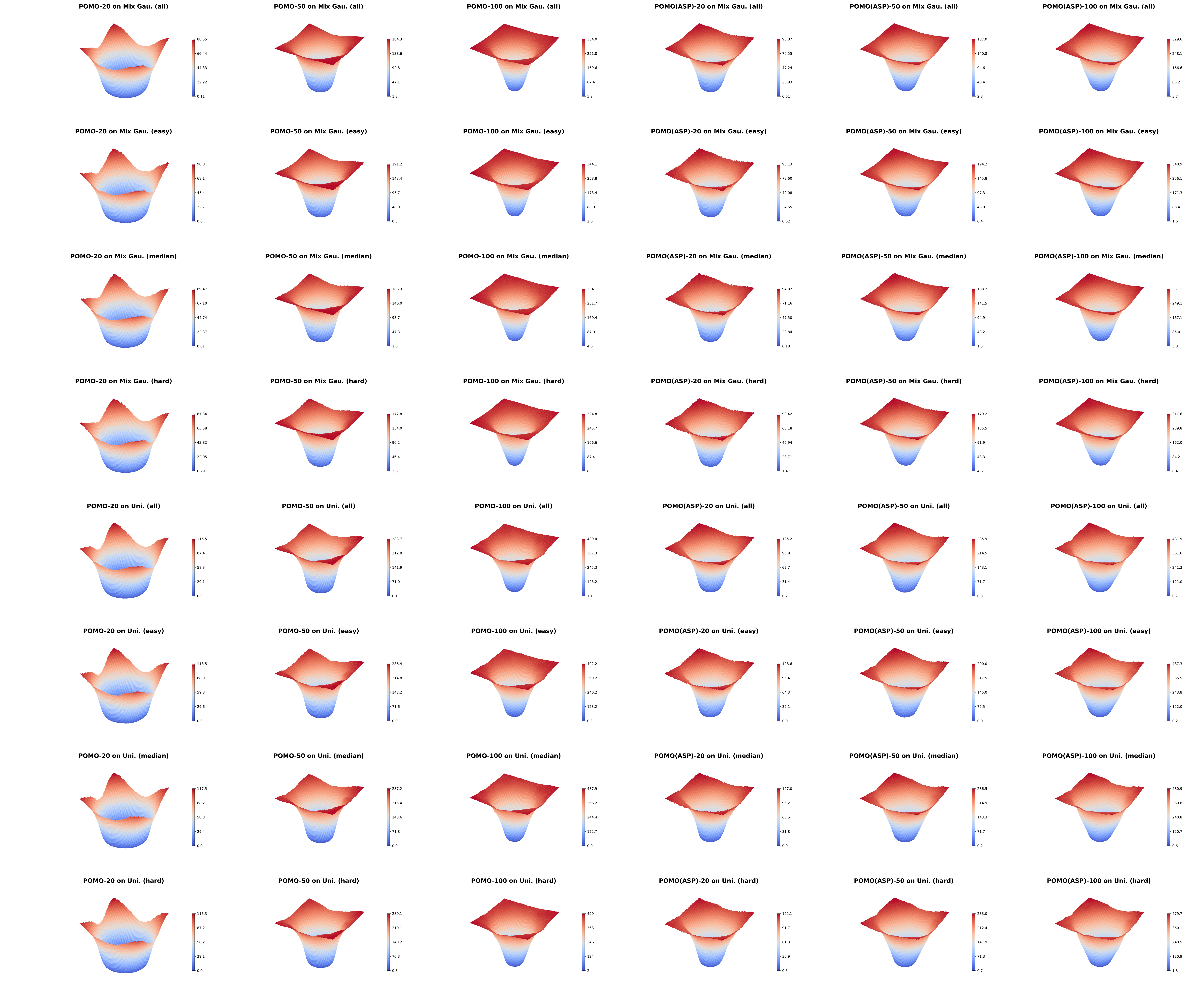}
    \caption{All the landscapes of optimality gap (\%) for POMO.}
    \label{fig:pomo landscape}
\end{figure}

We use the  “filter normalization” \cite{li2018visualizing} to visualize the objective function curvature of COP in Fig. \ref{fig:am landscape} and \ref{fig:pomo landscape} for AM and POMO. We evaluate 1000 instances for each set of permuted parameters, the range of permutation is from -1 to 1 and the step is 50. We evaluate AM and POMO with the combination of the following settings:
\begin{itemize}
    \item Different distributions: "Uni." and "Mix Gau." mean we plot the landscape of objective function evaluated on the instances sampling from uniform distribution and mix gaussian distribution
    \item Different training paradigm: "Neural solver(ASP)" means the results are obtained from the neural solver trained under the ASP framework, otherwise under the original setting
    \item Different problem scales: we evaluate neural solvers under different problem scales: 20, 50 and 100
    \item Different degrees of hardness: we separate the 1000 instances into three degrees of hardness according to the optimality gap: easy (instances of the first $\frac{1}{3}$ smallest optimality gap), median ($\frac{1}{3}\sim \frac{2}{3}$) and hard ($\frac{2}{3}\sim 1$). 
\end{itemize}
For example, "AM(ASP)-20 on Mix Gau. (all)" means: the landscape is obtained by evaluating AM trained from ASP framework on TSP20 and the 1000 instances are sampled from mix gaussian distribution.

\ifCLASSOPTIONcaptionsoff
\fi

\end{document}